\pdfoutput=1

\documentclass[11pt]{article}

\usepackage[final]{acl}

\usepackage{times}
\usepackage{latexsym}

\usepackage[T1]{fontenc}

\usepackage[utf8]{inputenc}

\usepackage{microtype}

\usepackage{inconsolata}

\usepackage{graphicx}
\usepackage{url}
\usepackage{amsmath, amsfonts}
\usepackage{color, xcolor}
\usepackage{multirow}
\usepackage{graphicx}
\usepackage{amssymb}
\usepackage{booktabs}
\usepackage{array}
\usepackage[normalem]{ulem}
\useunder{\uline}{\ul}{}
\usepackage{wrapfig}
\usepackage{hyperref}
\usepackage{float}

\title{Enhancing Unsupervised Sentence Embeddings via Knowledge-Driven Data
Augmentation and Gaussian-Decayed Contrastive Learning}

\author{
 \textbf{Peichao Lai\textsuperscript{1}},
 \textbf{Zhengfeng Zhang\textsuperscript{2}},
 \textbf{Wentao Zhang\textsuperscript{1}}, \\
 \textbf{Fangcheng Fu\textsuperscript{1}},
 \textbf{Bin Cui\textsuperscript{1}\thanks{indicates the corresponding author}}
 \\
 \textsuperscript{1}School of Computer Science, Peking University, \\
 \textsuperscript{2}College of Computer and Data Science, Fuzhou University \\
 \small{
  \href{mailto: bin.cui@pku.edu.cn}{bin.cui@pku.edu.cn}
 }
}

\begin{document}
    \maketitle
    \begin{abstract}
        Recently, using large language models (LLMs) for data augmentation has
        led to considerable improvements in unsupervised sentence embedding
        models. However, existing methods encounter two primary challenges: limited
        data diversity and high data noise. Current approaches often neglect
        fine-grained knowledge, such as entities and quantities, leading to
        insufficient diversity. Besides, unsupervised data frequently lacks
        discriminative information, and the generated synthetic samples may introduce
        noise. In this paper, we propose a pipeline-based data augmentation method
        via LLMs and introduce the Gaussian-decayed gradient-assisted Contrastive
        Sentence Embedding (GCSE) model\footnote{Code is available at: \url{https://github.com/aleversn/GCSE}} to enhance unsupervised sentence embeddings.
        To tackle the issue of low data diversity, our pipeline utilizes knowledge
        graphs (KGs) to extract entities and quantities, enabling LLMs to
        generate more diverse samples. To address high data noise, the GCSE
        model uses a Gaussian-decayed function to limit the impact of false hard
        negative samples, enhancing the model's discriminative capability.
        Experimental results show that our approach achieves state-of-the-art
        performance in semantic textual similarity (STS) tasks, using fewer data
        samples and smaller LLMs, demonstrating its efficiency and robustness
        across various models.
    \end{abstract}

    \section{Introduction}


    Sentence representation learning, a fundamental task in natural language processing (NLP), aims to generate accurate sentence embeddings to enhance performance in downstream tasks such as semantic inference \citep{reimers2019sentence}, retrieval \citep{thakur2021beir, wang-etal-2022-just}, and question answering \citep{DBLP:journals/pvldb/SenLQOEDSMSS20}. To improve computational efficiency and reduce labor costs, unsupervised sentence embedding methods based on contrastive learning \citep{gao2021simcse, wu-etal-2022-esimcse} have emerged as highly effective paradigms. Generally, contrastive learning operates on the principle that robust sentence embeddings should pull semantically similar sentences closer while pushing dissimilar ones further apart. The performance of unsupervised contrastive learning methods largely depends on the quantity and quality of training samples \citep{chen-etal-2022-generate}, highlighting the importance of strategies that effectively enhance both.


    Previous studies mainly focused on increasing the number of samples using
    rule-based word modifications \citep{wang2023sncse, wu-etal-2022-esimcse} or
    feature sampling and perturbation techniques \citep{10191335, chuang-etal-2022-diffcse}.
    Recent studies \citep{DBLP:conf/emnlp/ZhangLH23, DBLP:conf/naacl/WangLCSB24}
    use either few-shot manually constructed samples or zero-shot generalized
    refactoring instructions to create prompts that guide large language models
    (LLMs) in generating new samples from original sentences, increasing both the
    quantity and quality of the data. Although these methods have achieved
    commendable performance, two limitations remain:

    \textbf{Low Data Diversity.} Diverse data samples in sentence representation
    learning should contain varied expressions of the same knowledge. However,
    existing approaches often struggle to distinguish fine-grained semantic
    knowledge like entities and quantities in the context. Traditional methods modify
    sentences using limited patterns without considering fine-grained knowledge,
    restricting their effectiveness in enhancing sample diversity. Recent LLM-based
    methods like \citet{wang-etal-2024-improving-text}, SynCSE \citep{DBLP:conf/emnlp/ZhangLH23}
    and MultiCSR \citep{DBLP:conf/naacl/WangLCSB24}, adjust topic and entailment
    categories in prompts to guide the model in generating varied samples. These
    methods focus on the global context but lack precise control over the
    knowledge in the samples. Consequently, the diversity of generated samples
    is constrained by the probability distributions of LLMs, resulting in unpredictable
    data quality.

    \begin{figure}[t]
        \centering
        \includegraphics[width=0.85\columnwidth]{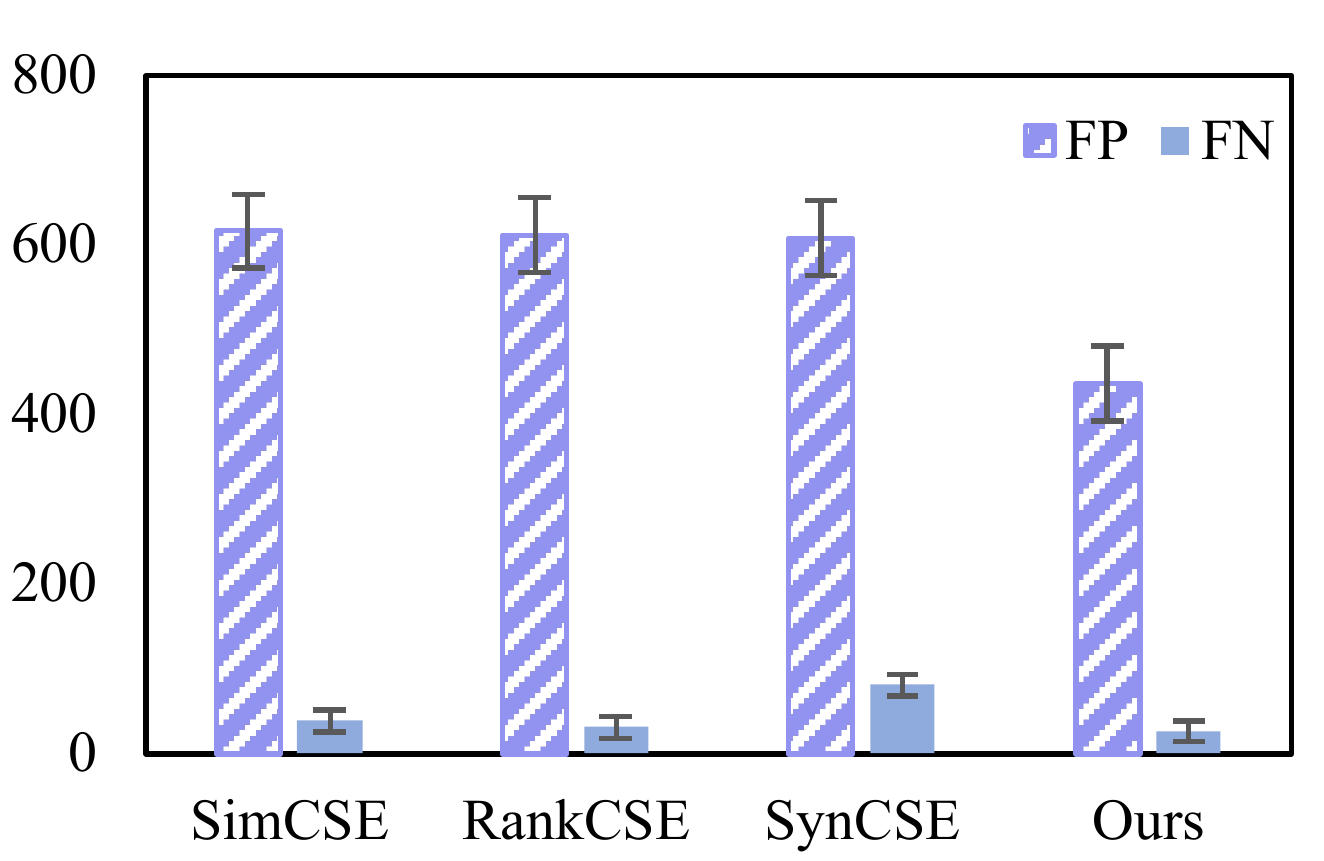}
        \caption{Comparison of false positives (FP) and negatives (FN). Both the
        predicted scores and labels are normalized (see details in Appendix~\ref{ap_snm}),
        where positives have a score greater than the label, while negatives lower
        than the label. False samples are identified when the root mean square
        error (RMSE) between the prediction and the label exceeds 0.2.}
        \label{f0}
    \end{figure}

    \textbf{High Data Noise.} Unsupervised sentence representation learning often
    suffers from data noise caused by confusing negative samples, which mainly
    arise from two sources. First, traditional methods generate datasets by
    duplicating samples to create positive instances, leading to negatives with similar
    surface-level semantics that affect the model's understanding of fine-grained semantic information \citep{DBLP:conf/cikm/MiaoDZ23, zhou-etal-2022-debiased}.
    Second, in data synthesis, differences in semantic distributions can cause the
    LLM's criteria for distinguishing between positive and negative samples to misalign
    with the target domain, introducing additional noise \citep{huang-etal-2023-adasent, poerner-schutze-2019-multi}.
    Existing method like MultiCSR attempts to remove noisy samples using linear
    programming, but this can eliminate potentially valuable samples and reduce data
    diversity. Figure~\ref{f0} compares various baselines on the STS-Benchmark
    development set. The results show that the prediction of false positives outnumber
    false negatives, and data synthesis in SynCSE increases false negatives, further
    supporting the above analysis.


    In this paper, we propose a pipeline-based data augmentation method using
    LLMs and introduce the Gaussian-decayed gradient-assisted Contrastive
    Sentence Embedding (GCSE) model to improve the performance of unsupervised
    sentence embedding methods. To address the issue of \textbf{\textit{low data
    diversity}}, we begin by extracting entities and quantities from the data samples
    and constructing a knowledge graph (KG). Next, we create
    a sentence construction prompt using the extracted knowledge to guide LLM in
    generating more diverse positive samples. To tackle \textbf{\textit{high
    data noise}}, we employ an evaluation model to annotate the synthesized data
    and initially filter out false positive samples. To further minimize the impact of
    false negatives while maintaining sample diversity, we align hard negatives with the evaluation model's distribution and reduce their gradient during the initial training step. Then, we leverage other
    in-batch negative samples to optimize the semantic space. Inspired by locally weighted linear regression \citep{DBLP:journals/air/AtkesonMS97}, we propose the GCSE model, which utilizes a Gaussian-decayed function to adjust prediction discrepancies between the GCSE model and the evaluation model. Initially, it reduces the gradient impact of hard negatives, gradually restoring their gradient weights as training progresses if they deviate significantly from the evaluation model's distribution. This approach prevents false negatives from being pushed further in the semantic space, promoting a more uniform distribution.

    \begin{table}[h]
        \centering
        \resizebox{\columnwidth}{!}{%
        \begin{tabular}{cccc}
            \toprule \textbf{Methods} & \textbf{Synthesis Approach} & \textbf{Use Knowledge} & \textbf{Denoise} \\
            \midrule SynCSE           & Few-shot Synthesis          & No                     & No               \\
            MultiCSR                  & Zero-shot Synthesis         & No                     & Yes              \\
            Ours                      & Zero-shot Synthesis         & Yes                    & Yes              \\
            \bottomrule
        \end{tabular}
        }
        \caption{Comparison of our methods and related LLM-based methods.}
        \label{t0}
    \end{table}

    \begin{figure*}[h!]
        \centering
        \includegraphics[width=\textwidth]{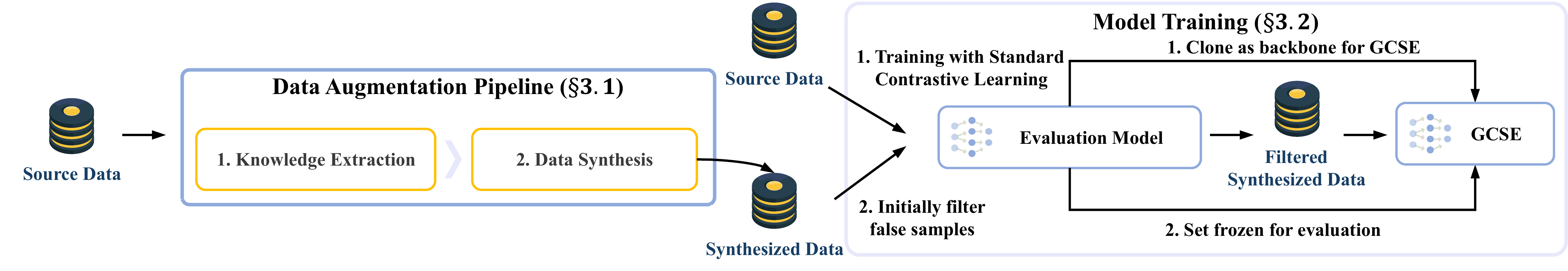}
        \caption{The overall workflow of our method.}
        \label{f6}
    \end{figure*}
    
    We highlight the key innovations of our approach in Table~\ref{t0}: (i) We are
    the first to incorporate fine-grained knowledge for sample synthesis in LLM-based
    methods. (ii) Unlike MultiCSR's denoising approach, our method retains more false
    samples for training rather than discarding them. (iii) Our data selection strategy is particularly well-suited for leveraging a local LLM to synthesize domain-specific samples from a limited number of samples, leading to improved performance. Experimental results demonstrate
    the efficiency of our model, outperforming previous best methods in average
    scores for semantic textual similarity (STS) tasks by 1.05\% with BERT-base,
    1.89\% with BERT-large, 0.50\% with RoBERTa-base, and 1.50\% with RoBERTa-large.

    In summary, our contributions are as follows: (1) \underline{\textit{New method.}}
    We introduce a pipeline-based data augmentation method using LLM for few-shot
    domain data and propose a Gaussian-decayed gradient-assisted Contrastive Sentence
    Embedding (GCSE) model to reduce data noise. (2) \underline{\textit{New perspective.}}
    To the best of our knowledge, we are the first to explore combining
    knowledge graphs with LLM to synthesize data, enhancing fine-grained sentence
    representation learning by generating diverse positive and negative samples.
    (3) \underline{\textit{State-of-the-art performance.}} Experimental results demonstrate
    that our method achieves superior performance on STS tasks while using fewer
    samples for data synthesis with smaller LLM parameters.





    \section{Related Work}

    Early work on sentence embeddings builds on the distributional hypothesis, predicting
    surrounding sentences \citep{NIPS2015_f442d33f, logeswaran2018an, hill-etal-2016-learning}
    or extending the word2vec framework \citep{DBLP:conf/nips/MikolovSCCD13} with
    n-gram embeddings \citep{pagliardini-etal-2018-unsupervised}. Post-processing
    techniques like BERT-flow \citep{li-etal-2020-sentence} and BERT-whitening
    \citep{DBLP:journals/corr/abs-2103-15316} address the anisotropy issue in pre-trained
    language models (PLMs), and more recent methods focus on generative
    approaches \citep{wang-etal-2021-tsdae-using, wu-zhao-2022-sentence} and regularizing
    embeddings to prevent representation degeneration \citep{DBLP:conf/emnlp/HuangTZLSGJD21}.
    Recently, contrastive learning approaches have become prominent, using various
    augmentation methods to derive different views of the same sentence \citep{zhang-etal-2020-unsupervised, giorgi-etal-2021-declutr, kim-etal-2021-self, gao2021simcse}.
    Among these, SimCSE uses dropout as a simple augmentation and achieves
    strong results in unsupervised STS tasks, inspiring further approaches like ArcCSE
    \citep{zhang-etal-2022-contrastive}, DiffCSE \citep{chuang-etal-2022-diffcse},
    GS-InfoNCE \citep{wu-etal-2022-smoothed}, and RankCSE \citep{DBLP:conf/acl/LiuLWWWX0C023}.

    With the advent of LLMs \citep{openai2023gpt4, qwen, DBLP:journals/corr/abs-2302-13971},
    some works attempt to utilize LLM for sentence representation learning. For example,
    \citet{DBLP:conf/acl/NiACMHCY22} uses T5 with mean pooling to obtain a
    sentence embedding model by fine-tuning on a large-scale NLI corpus; \citet{cheng-etal-2023-improving}
    uses prompt learning to measure the semantic similarity of sentence pairs;
    \citet{DBLP:journals/corr/abs-2402-15449} employs sentence repetition to
    enhance the capacity for sentence representation; AoE \citep{li-li-2024-aoe}
    optimize angle differences for improving supervised text embedding; and BeLLM
    \citep{li-li-2024-bellm} designs a Siamese structure for learning sentence
    embeddings.
    
    \section{Methodology}

    In this section, we present the data augmentation pipeline via LLM and the
    specific structure of the GCSE. As shown in Figure~\ref{f6}, we start by
    using a data augmentation pipeline to synthesize new samples from the source
    data, and then train our model with the filtered synthetic data.

    \subsection{Data Augmentation}
    \label{method_1}

\begin{figure}[t!]
        \centering
        \includegraphics[width=\columnwidth]{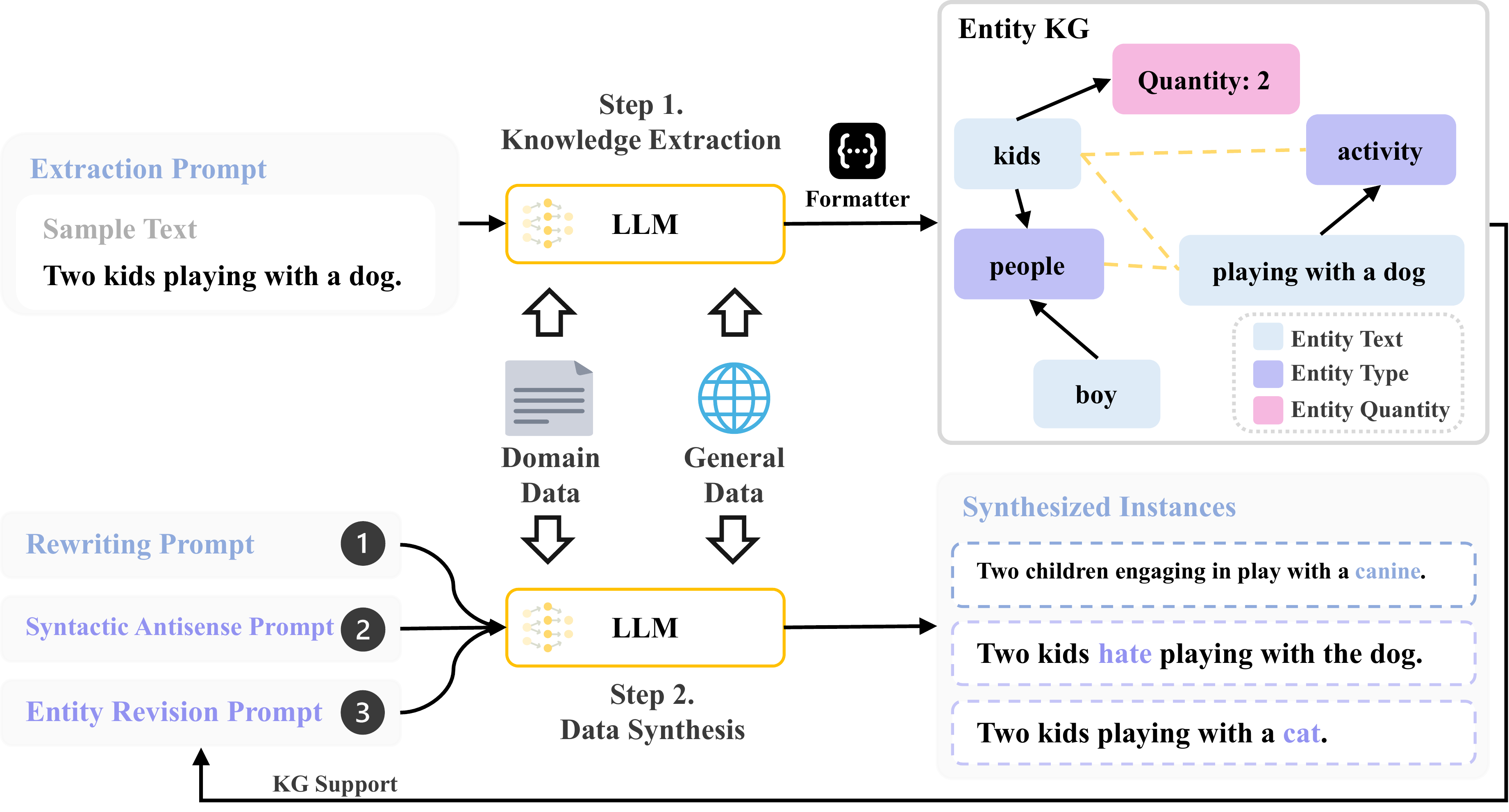}
        \caption{The pipeline of knowledge extraction and data synthesis, where the
        solid black arrows in the Entity KG are hard edges, and dotted yellow
        lines are soft edges.}
        \label{f1}
    \end{figure}

    In the data augmentation pipeline, we utilize both domain data and partial general
    data to balance domain-specific relevance and general-domain applicability.
    We start by extracting knowledge from the source data and then synthesize
    new data for our model training. The detailed structure of the pipeline is shown
    in Figure~\ref{f1}.

    \textbf{Knowledge Extraction and Integration.} The variety and relationships
    between samples directly impact model performance in sentence representation
    learning. A major challenge with existing LLM-based data synthesis methods is
    the limited diversity they generate for each short text. To trade off the
    low diversity of the generated samples with their relevance to the domain
    semantic space, we first design an extraction prompt to obtain entities and quantities
    from the given data. Formally, we denote the extraction prompt as $\mathcal{P}
    _{e}$, and LLM $\mathcal{L}$, suppose we finally extract instances with $d$ sample
    number, the knowledge set
    $\mathcal{K}_{i}=\left\{k_{i1}, \dots, k_{in}\right\}$ of each instance
    $x_{i}$ is computed in Equation~\ref{e1}, where $t_{j}$, $c_{j}$ and $q_{j}$
    represent the entity text, entity type, and quantity of $k_{i}$. $n$ is
    the size of $\mathcal{K}_{i}$, and $\mathcal{F}(\cdot)$ is the formatting function
    that converts text to a triplet. Next, we integrate all knowledge by establishing
    an entity knowledge graph $\mathcal{G}=\langle V,E\rangle$, where the node
    set $V$ contains all the $\langle t,c,q\rangle$ from $\mathcal{K}$:
    \begin{multline}
        \label{e1}\mathcal{K}=\bigcup_{i=1}^{d}{\mathcal{F}}([\mathcal{P}_{e};x_{i}
        ],{\mathcal{L}})=\\
        \bigcup_{i=1}^{d}\{\langle t_{i j},c_{i j},q_{i j}\rangle
        \mid j\in[1,n]\},
    \end{multline}
    \begin{equation}
        V=\{t_{i j},c_{i j},q_{i j}\mid i\in[1,d];j\in[1,n]\}.
    \end{equation}
    The edges $E$ consist of hard edges $E_{r}$ and soft edges $E_{s}$. As shown
    in Equations~\ref{e3} and \ref{e4}, $E_{r}$ represents the relationship
    between the entity text, type, and quantity of each $k\in\mathcal{K}$, and
    $E_{s}$ indicates the relationship between entity text in $k_{i j}$ and
    other entity text or type in the same instance $x_{i}$.
    \begin{equation}
        \resizebox{0.85\columnwidth}{!}{$\label{e3}E_{r}=\{(t_{i j},c_{i j})\cup(t_{i j},q_{i j})\mid
        i\in[1,d];j\in [1,n]\},$}
    \end{equation}
    \begin{equation}
        \resizebox{0.85\columnwidth}{!}{$\label{e4}E_{s}=\bigcup_{i=1}^{d}\{(t_{i j},t_{i k}),(t_{i j},c_{i l})\mid
        j,k,l\in[1,n];k,l\neq j\}.$}
    \end{equation}
    By defining hard and soft edges, we can more efficiently identify and replace
    entity nodes near the current node, improving the correlation between the
    synthesized instance and the source instance.

    \textbf{Data Synthesis via LLM.} Empirical evidence and model performance on
    standard datasets show that sentence embedding models struggle more with accurately
    identifying negative samples than positives \citep{chuang-etal-2022-diffcse, DBLP:conf/cikm/MiaoDZ23}.
    In the contrastive learning methods, the model acquires sentence embedding
    representation by calculating the distance between sentence-pairs. It aims to
    minimize the spatial distance between positive pairs and increase the spatial
    distance between negative pairs. Thus, it is essential to obtain negative samples
    that closely resemble the source instance in surface-level features, while
    positive samples should have diverse representations but still convey the
    same meaning as the source instance.

    In this study, we use LLM to generate positive samples through a rewrite prompt.
    We also focus on the impact of variations in entities and quantities within
    the samples. Negative samples are generated by the LLM at both the syntactic
    and fine-grained knowledge levels. The data synthesis prompts are divided
    into three main types: (1) Rewriting prompt, (2) Syntactic antisense prompt,
    and (3) Entity revision prompt. The first type is used to create positive
    samples, while the second and third types are used to create negative samples
    at the syntactic and knowledge levels, respectively.

    The ``rewriting prompt'' can be classified into three forms: directly
    requesting LLM to generate a new sentence instance using the ``rewrite'' instruction,
    creating the preceding part of the sentence instance, and generating based on
    the knowledge set of the instance. As the diversity of synthetic samples
    increases, the likelihood of generating false positives also rises. To
    address this, the next section involves scoring the generated samples using an
    evaluation model.
    
    The ``syntactic antisense prompt'' aims to modify the
    semantics to create a contradiction at the syntactic level. Such as transforming
    it into a positive or negative statement using explicit positive or negative words,
    or by expressing a contrary sentiment. This is an initial approach to synthesizing
    negative samples that preserves a strong coherence with the source instance in
    terms of sequence structure. However, it is deficient in generation diversity.
    To alleviate the issue, the ``entity revision prompt'' aims to enhance text
    diversity by replacing the entity text and quantity compared to the source
    instance. Simultaneously, to ensure the semantic relevance between the
    synthetic samples and the source instance, replacement entities are selected
    by searching for neighboring nodes on entity KG. We define
    $\mathcal{T(\cdot)}$ as the search function, and the replacement entity of $t
    _{i j}$ are computed as:
    \begin{equation}
        \mathcal{T}_{r}(t_{i j})\resizebox{0.7\columnwidth}{!}{$
            =\{t_{p}\mid(t_{i j},c_{i j})\in
        E_{r}\land (t_{p},c_{i j})\in E_{r}\},$}
    \end{equation}
    \begin{equation}
        \mathcal{T}_{s}(t_{i j})=\{t_{p}\mid(t_{i j},t_{p})\in E_{s}\},
    \end{equation}
    \begin{equation}
        \mathcal{T}_{p}(t_{i j})\resizebox{0.7\columnwidth}{!}{$
            =\{t_{p}\mid \exists t_{k}\in
        \mathcal{T}_{s}(t_{i j} )\cap \mathcal{T}_{s}(t_{p}) \land t_{p}\in \mathcal{T}_{r}(t_{i j})\},$}
    \end{equation}
    \begin{equation}
        \mathcal{T}(t_{i j})=\mathcal{T}_{r}(t_{i j})\cup \mathcal{T}_{p}(t_{i j}
        ),
    \end{equation}
    where the function $\mathcal{T}_{r}(\cdot)$ is used to search for entities that share a hard edge with the current entity, while $\mathcal{T}_{s}(\cdot)$ retrieves entities connected via a soft edge. The function $\mathcal{T}_{p}(\cdot)$ is designed to find a replacement entity $t_{p}$ that is of the same type as $t_{ij}$ and shares soft-edge connections with another in-context entity $t_k$. Finally, the replacement entity is randomly selected from the results of the search function $\mathcal{T}(t_{ij})$. Compared to random entity substitution, our strategy significantly improves the semantic relevance between the synthesized sample and the source instance.

    \subsection{Model Training}
    \label{general_cl}

    The training process of our model consists of two stages. First, we combine
    general and domain-specific data to train an evaluation model using standard
    unsupervised contrastive learning. This improves the uniformity of sentence embeddings
    in general scenarios and reduces the impact of semantic distribution limitations
    in the synthesized data, enhancing model robustness. Then, we freeze the
    evaluation model to filter synthetic data and help the GCSE model eliminate
    false hard negative sample noise.

    \textbf{General Contrastive Learning.} In the first stage, we follow the
    formulation of SimCSE \citep{gao2021simcse} to train the evaluation model.
    Formally, we define the encoder of the evaluation model as $E^{\prime}$,
    each unlabeled sentence instance as $x_{i}$, and its positive sample as $x_{i}
    ^{+}=x_{i}$. The representation of each instance is denoted as
    $\mathbf{h}^{\prime}=\mathcal{F}_{E^\prime}(x)$, the representations of $x_{i}$
    and $x_{i}^{+}$ are computed as $\mathbf{h}^{\prime}_{i}$ and $\mathbf{h}^{\prime
    +}_{i}$, respectively. Since the dropout mask in $E^{\prime}$ is random,
    $\mathbf{h}^{\prime}_{i}$ and $\mathbf{h}^{\prime +}_{i}$ are computed with
    the same input but with slightly different results. Then, the loss of
    evaluation model is defined as:
    \begin{equation}
        -\log\frac{e^{\text{sim}(\mathbf{h}^{\prime}_{i},\mathbf{h}^{\prime +}_{i})/\tau}}{\sum_{j=1}^{N}e^{\text{sim}(\mathbf{h}^{\prime}_{i},\mathbf{h}^{\prime +}_{j})/\tau}}
        ,
    \end{equation}
    where $N$ represents the size of each mini-batch, $\tau$ is a temperature hyperparameter,
    and $\text{sim}(\cdot)$ is the cosine similarity function.

    \begin{figure}[t!]
        \centering
        \includegraphics[width=\columnwidth]{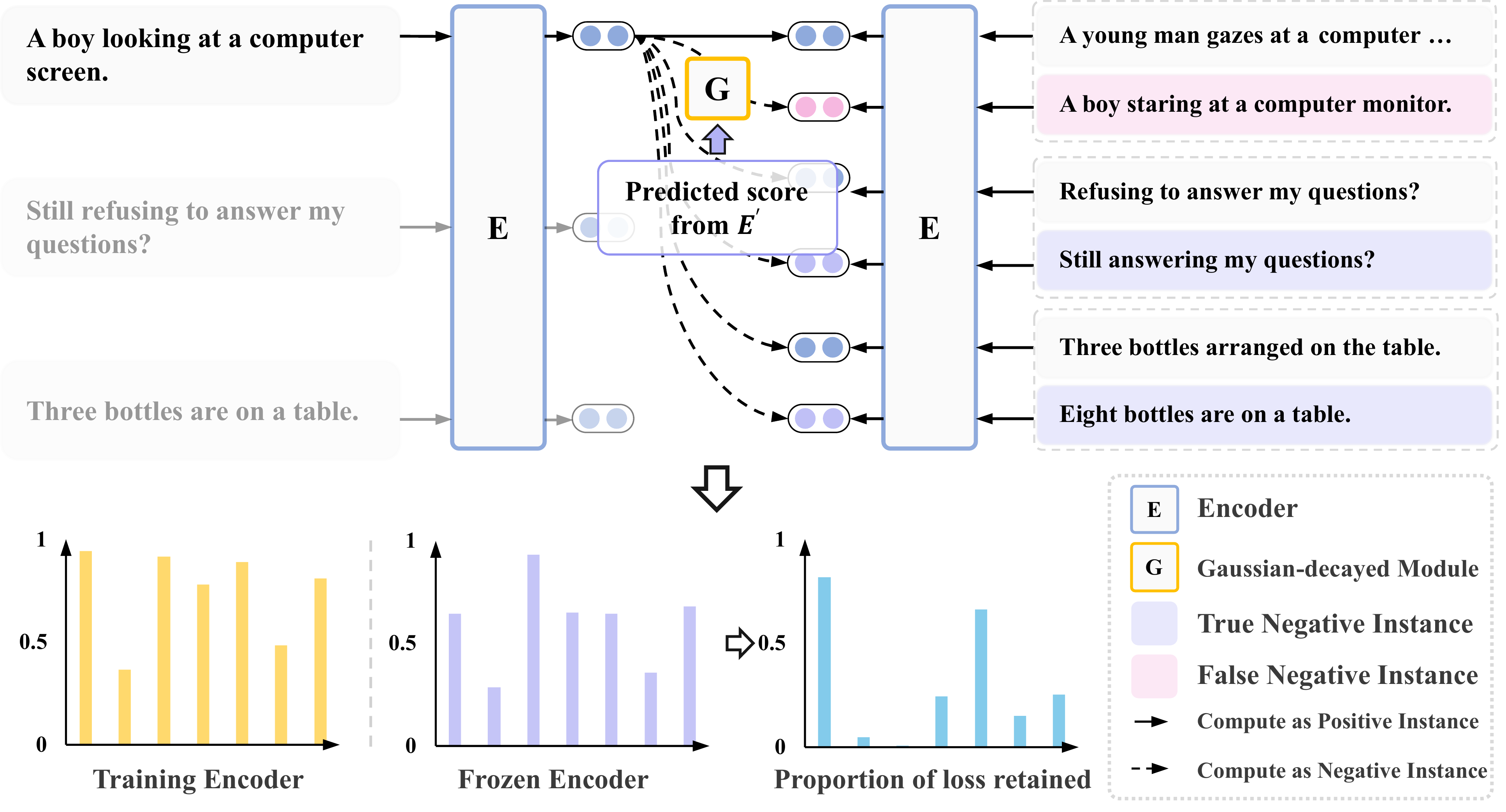}
        \caption{In-batch training with Gaussian-decayed on GCSE.}
        \label{f2}
    \end{figure}

    \textbf{Denoising Training.} In the second stage, we adopt a copy of the evaluation
    model as the backbone of GCSE and continue training on synthesized data. In
    this stage, each input is set as a triplet $(x_{i}, x_{i}^{+}, x_{i}^{-})$,
    where $x_{i}^{+}$ and $x_{i}^{-}$ stand for the positive and negative
    samples of $x_{i}$, respectively. Nevertheless, the synthesized data contains
    many potential false positive and false negative samples, necessitating the
    implementation of a filtering process. We use the frozen evaluation model to
    initially correct these inaccurate samples and build the ultimate triplet dataset.
    Let $\mathcal{S}(x_{i})=\left \{\hat{x}_{i 1}, \dots \hat{x}_{i m}\right\}$
    denotes the synthetic data set of $x_{i}$, where $m$ is the size of the set,
    and $x_{i}^{+}$, $x_{i}^{-}$ are calculated as:
    \begin{equation}
        x_{i}^{+}=
        \begin{cases}
            \hat{x}_{i j}, & \text{sim}(\mathbf{h}^{\prime}_{i}, \hat{\mathbf{h}}^{\prime}_{i j}) \geq \alpha,j \in [1,m] \\
            x_{i},         & \text{else}
        \end{cases},
    \end{equation}
    \begin{equation}
        x_{i}^{-}=
        \begin{cases}
            \hat{x}_{i j}, & \text{sim}(\mathbf{h}^{\prime}_{i}, \hat{\mathbf{h}}^{\prime}_{i j}) \leq \beta,j \in [1,m] \\
            x_{k},         & k\in[1,N], k \neq i
        \end{cases},
    \end{equation}
    where $\alpha$, $\beta$ are the threshold for positives and negatives, respectively.
    $x_{k}$ denotes a randomly selected instance from in-batch data. We can set a
    high value for $\alpha$ to reduce false positive samples. However, filtering
    out false negatives in synthetic data is more challenging. In theory, smaller
    $\beta$ can reduce more false negatives, but samples with low similarity to
    the source instance are easy to distinguish due to significant surface-level
    differences. As a result, training on these samples does not effectively
    improve the model's ability to distinguish fine-grained false positives. Therefore,
    we opt for a higher value of $\beta$. During training, we use a Gaussian-decayed
    function to align the distances of hard negative samples between the GCSE
    encoder $E$ and the frozen encoder $E^{\prime}$. As shown in Figure~\ref{f2},
    for each mini-batch of triplet inputs, both $E$ and $E^{\prime}$ compute similarity
    scores for the negative samples and their corresponding source instances.
    The loss for each instance in GCSE is defined as:
    \begin{equation}
        \resizebox{0.85\columnwidth}{!}{$-\log\frac{e^{\text{sim}({\bf h}_{i},{\bf h}_{i}^{+})/\tau}}{\sum_{j=1}^{N}e^{\text{sim}({\bf h}_{i},{\bf h}_{j}^{+})/\tau}+\sum_{\substack{j=1 \\ j \neq i}}^{N}e^{\text{sim}({\bf h}_{i},{\bf h}_{j}^{-})/\tau}+e^{G(s_{i},s^{\prime}_{i},\tau,\sigma)}},$
        }
    \end{equation}
    \begin{equation}
        \resizebox{0.8\columnwidth}{!}{$
            G(s_{i}, s'_{i}, \tau, \sigma)=\begin{cases}
                s_{i}\left(1 - e^{-\frac{(s_{i}- s'_{i})^{2}\tau^{2}}{2
            \, \sigma^{2}}}\right), &  s_{i} \leq s'_{i}\\
                s_{i},         & s_{i} > s'_{i}
            \end{cases},$}
    \end{equation}
    where $s_{i}=\text{sim}(\mathbf{h}_{i},\mathbf{h}^{-}_{i})$, $s^{\prime}_{i}=
    \text{sim}(\mathbf{h}^{\prime}_{i},\mathbf{h}^{\prime -}_{i})$. $G(\cdot)$
    is the Gaussian-decayed function, where the loss attenuation of the hard negative
    sample grows as the distance between $s_{i}$ and $s^{\prime}_{i}$ decreases,
    and $\sigma$ is a hyperparameter that controls the width of $G(\cdot)$. This
    implies that when $E$ initially calculates the hard negative sample, it
    follows the spatial distribution of $E^{\prime}$ as the ``established
    guidelines'' and uses other in-batch negative samples to further increase
    the spatial distance between negatives, effectively reducing the influence of
    false negatives. As training progresses, the spatial distribution of true hard
    negatives between $E$ and $E^{\prime}$ will progressively increase, and its
    gradient will be restored.

 \begin{table*}[h!]
        \centering
        \resizebox{0.8\textwidth}{!}{%
        \centering
        \begin{tabular}{m{3cm}<{\centering}|p{5.5cm}|cccccccc}
            \toprule \textbf{Model}                 & \textbf{Method}               & \textbf{STS-12} & \textbf{STS-13} & \textbf{STS-14} & \textbf{STS-15} & \textbf{STS-16} & \textbf{STS-B} & \textbf{SICK-R} & \textbf{Avg.}  \\
            \midrule \multirow{12}{*}{BERT-base}    & whitening\dag                 & 57.83           & 66.90           & 60.90           & 75.08           & 71.31           & 68.24          & 63.73           & 66.28          \\
                                                    & SimCSE\dag                    & 68.40           & 82.41           & 74.38           & 80.91           & 78.56           & 76.85          & 72.23           & 76.25          \\
                                                    & DiffCSE\dag                   & 72.28           & 84.43           & 76.47           & 83.90           & 80.54           & 80.59          & 71.23           & 78.49          \\
                                                    & PromptBERT$\clubsuit$         & 71.56           & 84.58           & 76.98           & 84.47           & 80.60           & 81.60          & 69.87           & 78.54          \\
                                                    & PCL$\spadesuit$               & 72.84           & 83.81           & 76.52           & 83.06           & 79.32           & 80.01          & 73.38           & 78.42          \\
                                                    & DebCSE\dag                    & 76.15           & 84.67           & 78.91           & {\ul 85.41}           & 80.55           & 82.99          & 73.60           & 80.33          \\
                                                    & RankCSE$\spadesuit$           & 75.66           & \textbf{86.27}  & 77.81           & 84.74           & 81.10           & 81.80          & 75.13           & 80.36          \\
                                                    & SynCSE (GPT-3.5 Turbo)*             & 75.86           & 82.19           & 78.71           & \textbf{85.63}  & 81.11           & 82.35          & {\ul 78.79}     & 80.66          \\
                                                    & MultiCSR (GPT-3.5 Turbo)$\clubsuit$ & 74.86           & 84.19           & 79.46           & 84.70           & 80.34           & {\ul 83.59}    & \textbf{79.37}  & 80.93          \\
                                                    & \textbf{GCSE (ChatGLM3-6B)}   & {\ul 78.14}  & 85.89  & 80.71  & 84.92  & 81.20  & 82.89 & 77.49  & 81.61 \\
                                                    & \textbf{GCSE (GLM4-9B-Chat)}       & 77.30  & {\ul 86.21}  & 80.60  & 84.98  & 81.48  & 83.22 & 77.82  & 81.66 \\
                                                    & \textbf{GCSE (Qwen2.5-32B-Instruct)}   & 77.83  & 86.07  & {\ul 80.77}  & 85.32  & {\ul 81.51}  & 83.26 & 78.17  & 81.85 \\
                                                    & \textbf{GCSE (GPT-3.5 Turbo)} & 77.88  & {\ul 86.21}  & \textbf{80.91}  & 84.98  & \textbf{81.60}  & 83.38 & 78.59  & {\ul 81.94} \\
                                                    & \textbf{GCSE (Deepseek-V3-0324)}   & \textbf{78.33}  & 86.12  & 80.31  & 85.32  & 81.38  & \textbf{83.62} & {\ul 78.79}  & \textbf{81.98} \\
            \midrule \multirow{8}{*}{BERT-large}    & SimCSE\dag                    & 70.88           & 84.16           & 76.43           & 84.50           & 79.76           & 79.26          & 73.88           & 78.41          \\
                                                    & PCL$\spadesuit$               & 74.87           & 86.11           & 78.29           & 85.65           & 80.52           & 81.62          & 73.94           & 80.14          \\
                                                    & DebCSE\dag                    & 76.82           & 86.36           & 79.81           & 85.80     & 80.83           & 83.45          & 74.67           & 81.11          \\
                                                    & RankCSE$\spadesuit$           & 75.48           & 86.50           & 78.60           & 85.45           & 81.09           & 81.58          & 75.53           & 80.60          \\
                                                    & SynCSE (GPT-3.5 Turbo)*             & 74.24           & 85.31           & 79.41           & 85.71           & 81.76  & 82.61          & 79.25     & 81.18          \\
                                                    & \textbf{GCSE (ChatGLM3-6B)}   & 77.69  & 86.98  & 81.68  & 86.01  & 81.89  & 84.28 & 79.43  & 82.57 \\
                                                    & \textbf{GCSE (GLM4-9B-Chat)}       & 78.17  & 87.02  & {\ul 82.08}  & \textbf{86.62}  & 82.04  & 84.71 & 79.53  & 82.89 \\
                                                    & \textbf{GCSE (Qwen2.5-32B-Instruct)}   & {\ul 78.34}  & 87.02  & 81.88  & {\ul 86.39}  & {\ul 82.29}  & {\ul 84.80} & 79.97  & 82.96 \\
                                                    & \textbf{GCSE (GPT-3.5 Turbo)} & \textbf{78.60}  & \textbf{87.27}  & 82.18  & 85.90  & \textbf{82.30}  & 84.77 & {\ul 80.09}  & {\ul 83.02} \\
                                                    & \textbf{GCSE (Deepseek-V3-0324)}   & 78.11  & {\ul 87.22}  & \textbf{82.23}  & 86.31  & 82.13  & \textbf{84.93} & \textbf{80.55}  & \textbf{83.07} \\
            \midrule \multirow{12}{*}{RoBERTa-base} & whitening\dag                 & 46.99           & 63.24           & 57.23           & 71.36           & 68.99           & 61.36          & 62.91           & 61.73          \\
                                                    & SimCSE\dag                    & 70.16           & 81.77           & 73.24           & 81.36           & 80.65           & 80.22          & 68.56           & 76.57          \\
                                                    & DiffCSE\dag                   & 70.05           & 83.43           & 75.49           & 82.81           & 82.12           & 82.38          & 71.19           & 78.21          \\
                                                    & PromptRoBERTa$\clubsuit$      & 73.94           & 84.74           & 77.28           & 84.99           & 81.74           & 81.88          & 69.50           & 79.15          \\
                                                    & PCL$\spadesuit$               & 71.13           & 82.38           & 75.40           & 83.07           & 81.98           & 81.63          & 69.72           & 77.90          \\
                                                    & DebCSE\dag                    & 74.29           & 85.54     & 79.46           & 85.68     & 81.20           & 83.96          & 74.04           & 80.60          \\
                                                    & RankCSE$\spadesuit$           & 73.20           & \textbf{85.95}  & 77.17           & 84.82           & 82.58           & 83.08          & 71.88           & 79.81          \\
                                                    & SynCSE (GPT-3.5 Turbo)$\dag\dag$    & 74.61           & 83.76           & 77.89           & 85.09           & 82.28           & 82.71          & 78.88           & 80.75          \\
                                                    & MultiCSR (GPT-3.5 Turbo)$\clubsuit$ & 75.61           & 84.33           & 80.10           & 84.98           & 82.13           & 84.54    & \textbf{79.67}  & 81.62          \\
                                                    & \textbf{GCSE (ChatGLM3-6B)}   & 76.95  & {\ul 85.59}  & 80.43  & 85.90  & \textbf{83.20}  & \textbf{84.62} & 77.28  & 82.00 \\
                                                    & \textbf{GCSE (GLM4-9B-Chat)}       & {\ul 77.83}  & 84.62  & 80.17  & {\ul 86.21}  & 82.99  & 84.05 & 78.33  & 82.03 \\
                                                    & \textbf{GCSE (Qwen2.5-32B-Instruct)}   & 77.81  & 84.56  & 80.23  & 86.13  & {\ul 83.19}  & 84.38 & 78.06  & 82.05 \\
                                                    & \textbf{GCSE (GPT-3.5 Turbo)} & \textbf{78.03}  & 83.79  & \textbf{80.61}  & \textbf{86.28}  & 82.76  & 84.31 & {\ul 79.01}  & {\ul 82.11} \\
                                                    & \textbf{GCSE (Deepseek-V3-0324)}   & 77.77  & 84.33  & {\ul 80.60}  & 86.01  & 82.75  & {\ul 84.60} & 78.77  & \textbf{82.12} \\
            \midrule \multirow{8}{*}{RoBERTa-large} & SimCSE\dag                    & 72.86           & 83.99           & 75.62           & 84.77           & 81.80           & 81.98          & 71.26           & 78.90          \\
                                                    & PCL$\spadesuit$               & 74.08           & 84.36           & 76.42           & 85.49           & 81.76           & 82.79          & 71.51           & 79.49          \\
                                                    & DebCSE\dag                    & 77.68           & 87.17           & 80.53           & 85.90           & 83.57           & 85.36          & 73.89           & 82.01          \\
                                                    & RankCSE$\spadesuit$           & 73.20           & 85.83           & 78.00           & 85.63           & 82.67           & 84.19          & 73.64           & 80.45          \\
                                                    & SynCSE (GPT-3.5 Turbo)$\dag\dag$    & 75.45           & 85.01           & 80.28           & 86.55           & 83.95           & 84.49          & \textbf{80.61}  & 82.33          \\
                                                    & \textbf{GCSE (ChatGLM3-6B)}   & 76.10  & 86.64  & 81.21  & 85.90  & 83.99  & 85.51 & 79.11  & 82.64 \\
                                                    & \textbf{GCSE (GLM4-9B-Chat)}       & 77.94  & 87.00  & 82.34  & 86.52  & 84.27  & 86.19 & 78.38  & 83.23 \\
                                                    & \textbf{GCSE (Qwen2.5-32B-Instruct)}   & 77.79  & {\ul 87.45}  & 82.22  & {\ul 87.86}  & \textbf{84.62}  & \textbf{86.75} & 78.30  & 83.57 \\
                                                    & \textbf{GCSE (GPT-3.5 Turbo)} & \textbf{78.21}  & \textbf{87.47}  & \textbf{82.76}  & 87.79  & {\ul 84.40}  & 86.15 & 80.02  & \textbf{83.83} \\
                                                    & \textbf{GCSE (Deepseek-V3-0324)}   & {\ul 78.11}  & 87.40  & {\ul 82.61}  & \textbf{88.00}  & 84.15  & {\ul 86.32} & {\ul 80.12}  & {\ul 83.82} \\
            \bottomrule
        \end{tabular}%
        }
        \caption{Comparison of Spearman's correlation results on STS tasks. The values in parentheses indicate using data synthesized by different LLMs. The values in bold and underlined indicate the best and second-best values, respectively. ``\dag'': results from
        \citet{DBLP:conf/cikm/MiaoDZ23}, ``$\clubsuit$'': results from
        \citet{DBLP:conf/naacl/WangLCSB24}, ``$\spadesuit$'': results from
        \citet{DBLP:conf/acl/LiuLWWWX0C023}, ``$\dag\dag$'': results from
        \citet{DBLP:conf/emnlp/ZhangLH23}. ``*'': we reproduce the results with
        the officially released corpus from \citet{DBLP:conf/emnlp/ZhangLH23}. GCSE
        has significant differences with all comparable baselines on the t-test ($p
        \textless 0.5\%$).}
        \label{t1}
    \end{table*}

    \begin{table}[t!]
    \centering
    \resizebox{0.8\columnwidth}{!}{%
    \begin{tabular}{clc}
    \toprule
    \textbf{Model}                 & \textbf{Method}      & \textbf{Avg.}  \\ \midrule
    \multirow{3}{*}{BERT-base}     & GCSE (ChatGLM3-6B)   & 81.35          \\
                                   & GCSE (GLM4-9B-Chat)  & 81.58          \\
                                   & GCSE (GPT-3.5 Turbo) & \textbf{81.92} \\ \midrule
    \multirow{3}{*}{BERT-large}    & GCSE (ChatGLM3-6B)   & 81.49          \\
                                   & GCSE (GLM4-9B-Chat)  & 82.05          \\
                                   & GCSE (GPT-3.5 Turbo) & \textbf{82.76} \\ \midrule
    \multirow{3}{*}{Roberta-base}  & GCSE (ChatGLM3-6B)   & 81.23          \\
                                   & GCSE (GLM4-9B-Chat)  & 81.70          \\
                                   & GCSE (GPT-3.5 Turbo) & \textbf{82.18} \\ \midrule
    \multirow{3}{*}{Roberta-large} & GCSE (ChatGLM3-6B)   & 82.62          \\
                                   & GCSE (GLM4-9B-Chat)  & 82.91          \\
                                   & GCSE (GPT-3.5 Turbo) & \textbf{83.83} \\ \bottomrule
    \end{tabular}%
    }
    \caption{Spearman's correlation results on STS tasks under the low-resource domain data setting. The best results are highlighted in bold.}
    \label{t10}
    \end{table}

    \section{Experiment}

    \subsection{Experiment Setup}
    \label{exp_setup}

    \textbf{Training:} In our main experiments, we evaluate model performance under two settings: (1) the default setting, where samples are synthesized using Wikipedia texts following \citet{gao2021simcse}; and (2) a simulated low-resource, high-quality setup using a smaller yet diverse set of domain-specific and general data. Specifically, we use a subset of the NLI dataset from \citet{gao2021simcse} as the general data, and select the training sets from STS-12 \citep{DBLP:journals/corr/abs-1708-00055} (2.2k samples), PAWS \citep{DBLP:conf/naacl/ZhangBH19} (3.5k samples), and SICK \citep{DBLP:conf/lrec/MarelliMBBBZ14} (4.5k samples) as domain-specific data. To simulate an unsupervised learning scenario, we only include the unlabeled portions of these datasets. In this experiment, the sample ratio between domain and general data is set to 1:3.
    
    We adopt ChatGLM3-6B
    \citep{glm2024chatglm}, GLM4-9B-Chat \citep{glm2024chatglm}, Qwen2.5-32B-Instruct \citep{qwen2.5, qwen2}, GPT-3.5 Turbo \citep{openai_chatgpt_2022} and Deepseek-V3-0324 \citep{deepseekai2024deepseekv3technicalreport}
    as LLMs for data synthesis, respectively. We choose BERT \citep{devlin-etal-2019-bert}
    and RoBERTa \citep{DBLP:journals/corr/abs-1907-11692} as the backbone models
    of GCSE. In the stage of Gaussian-decayed training on synthesized data, the
    filtering thresholds of $\alpha$ and $\beta$ are set as 0.9 and 0.75, respectively.
    The temperature of $\tau$ is set as 0.05, and the $\sigma$ of $G(\cdot)$ is set
    as 0.01. In the first stage training, the evaluation model is firstly trained
    on the unlabeled dataset of all general data and domain data. One copy
    instance of the evaluation model is then utilized as the pre-trained model
    for GCSE, while the original instance is set to be frozen to filter synthesized
    data and provide guidance for GCSE. In the second stage, GCSE is trained on the
    filtered synthesized data, and the sentence embedding is obtained from the
    last output hidden states of the first token. During the data augmentation phase, we used an NVIDIA A800 80G for LLM-based data synthesis. In the training phase, we conducted training and validation on eight NVIDIA TITAN RTX GPUs.

    \textbf{Evaluation:} To validate our method for sentence embeddings, we
    evaluated the model's performance on semantic textual similarity (STS) tasks,
    we use the standard evaluation method, measuring model performance with
    Spearman's correlation, and we adopt SentEval\footnote{https://github.com/facebookresearch/SentEval}
    \citep{DBLP:conf/lrec/ConneauK18} as the evaluation tool, which contains
    seven STS subsets: STS 2012-2016 \citep{agirre-etal-2012-semeval, agirre-etal-2013-semeval, agirre-etal-2014-semeval, agirre-etal-2015-semeval, agirre-etal-2016-semeval},
    the STS-Benchmark \citep{DBLP:journals/corr/abs-1708-00055} and the SICK
    Relatedness \citep{DBLP:conf/lrec/MarelliMBBBZ14}. Additionally, we compared
    the reranking task performance on Appendix~\ref{a_rk}, and the performance
    of our model with other methods on transfer tasks in SentEval to evaluate
    its applicability in Appendix~\ref{a_trans}.

    \textbf{Baselines:} We compare our method with mainstream unsupervised
    sentence embedding baselines: BERT-whitening \citep{DBLP:journals/corr/abs-2103-15316},
    SimCSE \citep{gao2021simcse}, DiffCSE \citep{DBLP:conf/naacl/ChuangDLZCS0YKG22},
    PromptBERT \citep{jiang-etal-2022-promptbert}, PCL \citep{wu-etal-2022-pcl},
    CARDS \citep{10.1145/3477495.3531823}, DebCSE \citep{DBLP:conf/cikm/MiaoDZ23}
    and RankCSE \citep{DBLP:conf/acl/LiuLWWWX0C023}. In addition, we further compare
    two baselines: SynCSE \citep{DBLP:conf/emnlp/ZhangLH23} and MultiCSR \citep{DBLP:conf/naacl/WangLCSB24},
    which use LLM for data synthesizing in whole NLI datasets. To verify the
    effectiveness of our data synthesis method, we choose their results of using
    GPT-3.5 Turbo for comparison.
    
    \subsection{Main Results}

    \textbf{STS Tasks:} The overall results of the STS tasks are shown in Table~\ref{t1}.
    Our approach, utilizing synthetic samples from Deepseek-V3-0324 and GPT-3.5 Turbo achieve the best performance across all backbones when compared to other unsupervised
    baselines. Even with synthetic samples from ChatGLM3-6B, our method still
    outperforms previous approaches on all backbones.
    This highlights the applicability of our method, as it can be effectively
    applied to multiple models. Compared to the standard unsupervised SimCSE,
    Spearman's correlation of GCSE (ChatGLM3-6B) is improved by an average of
    5.40\% on the base models and 3.95\% on the large models. On the strong baseline
    RankCSE, GCSE (ChatGLM3-6B) achieved a 1.90\% improvement over its average performance,
    demonstrating the effectiveness of the LLM data synthesis process.
    
    Furthermore, compared to the two state-of-the-art baseline models SynCSE and MultiCSR, both of which rely on LLMs for data synthesis, our approach consistently achieves better performance across all backbone models. Table~\ref{t10} further reports our results under a simulated low-resource, high-quality domain-specific setting. The results show that our method, which utilizes local LLMs, achieves higher average Spearman correlations than the GPT-3.5 Turbo-based versions of both baseline models. It is also worth noting that our method uses only 14\% of the sample size compared to SynCSE and MultiCSR, which rely on the full NLI datasets. These results demonstrate the effectiveness of our data synthesis method and our domain-oriented sample selection strategy.

    \subsection{Analysis}

    \begin{table*}[h!]
        \centering
        \resizebox{0.8\textwidth}{!}{%
        \begin{tabular}{p{5.5cm}cccccccc}
            \toprule \textbf{Method}    & \textbf{STS-12} & \textbf{STS-13} & \textbf{STS-14} & \textbf{STS-15} & \textbf{STS-16} & \textbf{STS-B} & \textbf{SICK-R} & \textbf{Avg.}  \\
            \midrule GCSE (ChatGLM3-6B) & \textbf{76.91}  & 85.48  & \textbf{79.49}  & \textbf{84.28}           & \textbf{82.65}  & \textbf{83.90}          & \textbf{76.72}           & \textbf{81.35} \\
            \hspace{1em}w/o stage-2     & 71.85           & 83.65           & 76.84           & 83.37           & 78.74           & 79.10          & 71.69           & 77.89          \\
            \hspace{1em}w randomly      & 71.94           & 84.03           & 76.99           & 83.65           & 79.11           & 78.66          & 69.28           & 77.67          \\
            \hspace{1em}w/o filtering   & 74.65           & 83.54           & 77.39           & 83.27           & 79.97           & 79.66          & 74.27           & 78.96          \\
            \hspace{1em}w/o decay       & 76.26           & \textbf{85.98}           & 79.35           & 84.09  & 82.12           & 83.85 & 76.00           & 81.09          \\
            \hspace{1em}w/o general     & 75.44           & 85.55           & 79.19           & 84.91           & 80.23           & 81.57          & 74.14           & 80.15          \\
            \hspace{1em}w/o domain      & 75.59           & 85.66           & 78.93           & 84.09           & 80.87           & 82.29          & 76.00  & 80.49          \\
            \bottomrule
        \end{tabular}%
        }
        \caption{Ablation studies of STS tasks on BERT-base. Other PLMs yield similar
        patterns to BERT-base.}
        \label{t3}
    \end{table*}

    \textbf{Ablation Studies:} We analyze the impact of each module or strategy in
    GCSE (ChatGLM3-6B) under domain-specific setting and report the results in Table~\ref{t3}. First, ``w/o stage-2'' refers
    to the results obtained without training in the second stage. This leads to
    a significant decrease in performance compared to the default model, which is
    the performance of the evaluation model and is similar to the conventional unsupervised
    SimCSE. Then, ``w randomly'' refers to the direct use of the instance itself
    as a positive sample in the combination dataset of domain and general data,
    while randomly selecting a negative instance from the dataset. We can observe
    that its performance in this case is even worse than the evaluation model.
    This demonstrates that the diversity of positive samples and the quality of
    negative samples significantly impact the performance of the model. ``w/o filtering''
    indicates the results of training by skipping evaluation model filtering and
    directly using the data synthesized by LLM. The results show that the
    performance of the model is significantly affected when false positive and
    negative samples are introduced without filtering. We investigate the impact
    of the Gaussian-decayed function by removing it, and the results are shown
    in ``w/o decay''. We can observe that the default model performs better overall
    than when the Gaussian-decayed function is removed, indicating that it can
    filter out potential false negative sample noise. Finally, we analyze the
    necessity of including general data and domain data in ``w/o general'' and ``w/o
    domain'' respectively. It can be observed that removing either of them results
    in a decline in performance, which indicates the significance of domain data
    and the essentiality of general data in our method.

    \begin{table}[h!]
        \centering
        \resizebox{0.7\columnwidth}{!}{%
        \begin{tabular}{cc}
            \toprule \textbf{Method} & \textbf{Spearman's} \\
            \midrule unsup-SimCSE    & 75.59               \\
            RankCSE                  & 79.74               \\
            SynCSE (GPT-3.5 Turbo)         & 91.58               \\
            GCSE (ChatGLM3-6B)       & \textbf{93.77}      \\
            \bottomrule
        \end{tabular}
        }
        \caption{Comparison of Spearman's correlation results on the synthetic
        data of the STS-Benchmark development set.}
        \label{t4}
    \end{table}

    \textbf{Analysis of entities and quantities awareness:} We analyze GCSE
    awareness of entities and quantities by constructing a dataset using the
    data synthesis method in Section~\ref{method_1} on the STS-Benchmark
    development set. Then, the similarity scores of each triplet in the dataset
    are annotated by two supervised pre-trained models: ``sup-simcse-bert-large''
    and ``sup-simcse-roberta-large''. The final label is the average score of
    the similarity calculated by both models. We evaluate Spearman's correlation
    scores of GCSE and the other three strong baselines on the backbone of the BERT-base
    model, and the results are shown in Table~\ref{t4}. Our GCSE achieves the best
    result and outperforms RankCSE by 14.03\%. In this case, both SynCSE and
    GCSE achieve significant improvements over methods without LLM. This might be
    due to the similarity of the semantic representation space between the training
    set and the development set, both of which are synthesized via LLM. Nevertheless,
    GCSE shows a notable enhancement in performance of 2.19\% compared to SynCSE,
    demonstrating that its understanding of the entities and quantities in
    sentences has enhanced to a certain degree.

    \begin{figure*}[h!] 
        \centering
        \begin{minipage}{0.3\textwidth} 
            \centering
            \includegraphics[width=\linewidth]{
                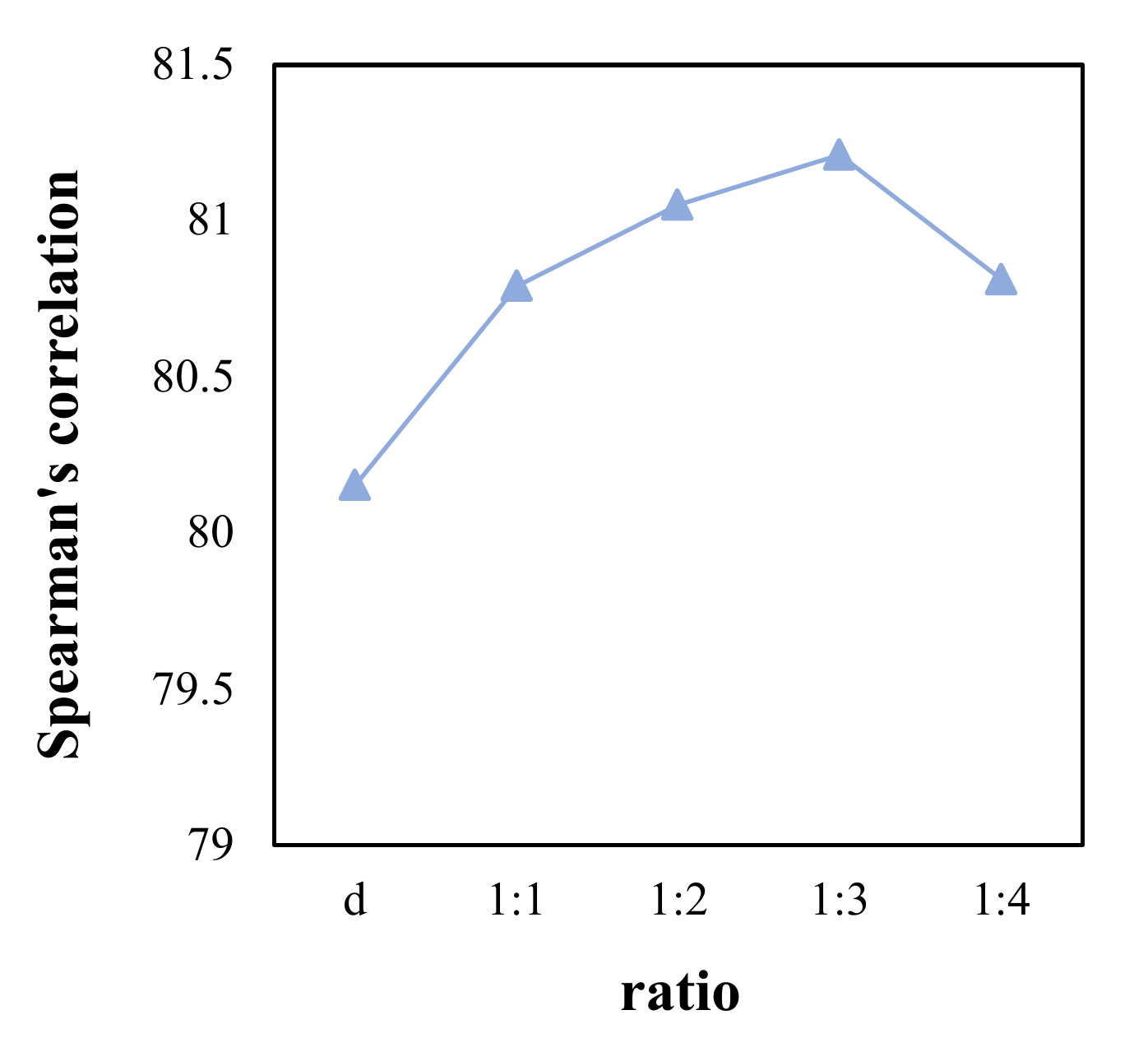
            } 
            \caption{Spearman's correlation against the ratio of domain data to
            general data on the STS tasks.}
            \label{f4}
        \end{minipage}
        \hfill 
        \begin{minipage}{0.635\textwidth} 
            \centering
            \includegraphics[width=\linewidth]{
                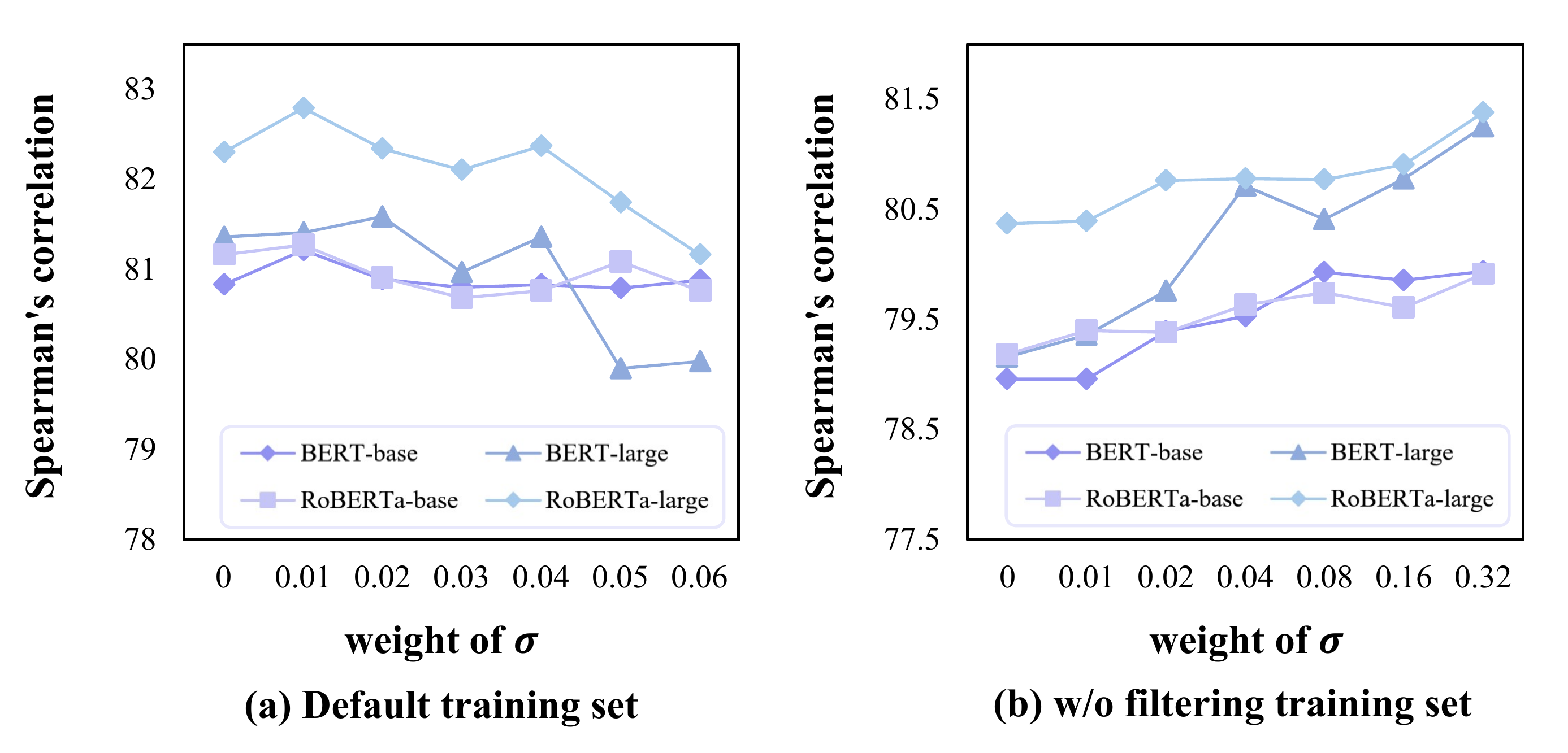
            } 
            \caption{Spearman's correlation against the weight of the Gaussian-decayed
            on the STS tasks.}
            \label{f3}
        \end{minipage}
    \end{figure*}

    \subsection{Impact on the ratio between domain and general data} Figure~\ref{f4}
    presents the trend of the GCSE Spearman's correlation result as the proportion
    of general data introduced increases, where ``d'' represents that only using
    the domain data. The results show that adding a certain amount of general data
    improves performance on STS tasks. However, when the size of general data exceeds
    three times that of domain data, performance starts to decline. This suggests
    that incorporating a moderate amount of external data enhances the uniformity
    of sentence embeddings. But as the out-of-domain data grows, the influence of
    domain-specific data on training weakens. Overall, the results indicate that
    domain data improves the model's ability to represent target domain sentences,
    while general data helps with sentence embedding uniformity.

    \begin{figure*}[h!]
        \centering
        \includegraphics[width=\textwidth]{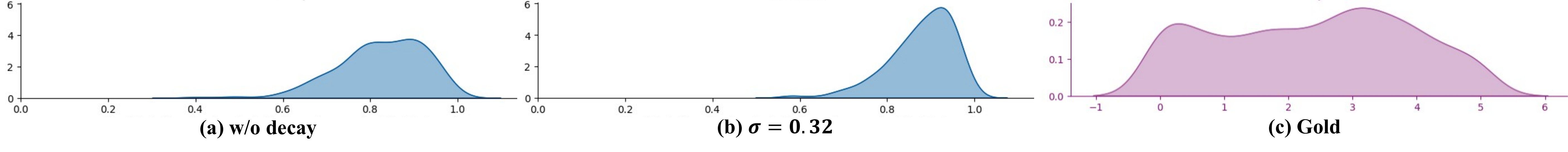}
        \caption{Density plots of the STS-Benchmark development set with labels
        $\geq 4$, which is evaluated by GCSE (ChatGLM3-6B) in domain-specific setting with different $\sigma$ weights. (c)
        is the density plot of gold labels.}
        \label{f5}
    \end{figure*}

    \subsection{Impact of the Gaussian-decayed}\label{a_i_gau} To further investigate the effectiveness
    of the Gaussian-decayed function, we analyze the GCSE (ChatGLM3-6B) performance in domain-specific setting against
    the weight of $\sigma$ on the synthesized data, both with and without filtering.
    As shown in Figure~\ref{f3}, we use the synthesized data without filtering to
    evaluate the efficacy of the Gaussian-decayed function in eliminating false negative
    samples, and results are presented in Figure~\ref{f3} (b). It is clear that the
    model's performance improves as the weight of $\sigma$ grows. This suggests
    that a greater $\sigma$ weight enhances the model's effectiveness in
    mitigating the impact of false negative samples. It is important to acknowledge
    that a higher $\sigma$ does not necessarily indicate better performance. As
    shown in Figure~\ref{f3} (a), an increase in $\sigma$ at the initial stage contributes
    to enhancing the model's performance. Nevertheless, as the weight of $\sigma$
    increases, the performance of backbones generally declines, resulting in the
    model adhering too strictly to the ``established guidelines''. Consequently,
    it impacts the efficacy of learning from the hard negative samples. We
    further use the density plots to visualize the prediction on the STS-Benchmark
    development set in Figure~\ref{f5}. These models are trained on the synthesized
    data without filtering. We can observe that in Figure~\ref{f5} (a), the distribution
    of prediction results for labels $\geq 4$ is significantly shifted to the left.
    Compared with the results in Figure~\ref{f5} (b), this issue is effectively alleviated,
    demonstrating the effectiveness of the Gaussian-decayed function in reducing
    the influence of false negative samples. To further verify the applicability
    of the Gaussian-decayed function, we applied it to SynCSE and verified the
    performance in Appendix~\ref{a_gau}.

    \section{Conclusion}

    In this paper, we propose a pipeline-based data augmentation method using
    LLM to enhance data diversity in sentence representation learning. By leveraging
    knowledge of entities and quantities, our approach improves the model's
    ability to capture fine-grained semantic distinctions. The Gaussian-decayed function
    in our GCSE model further reduces noise in the generated data. Extensive
    experiments on STS and reranking tasks show that our method achieves state-of-the-art
    results with fewer synthesized samples and a more lightweight LLM, demonstrating
    its effectiveness and efficiency.

    \section*{Limitations}
    While our data augmentation method achieves promising results across LLMs with varying parameter scales, we observe performance discrepancies depending on the size of the LLM used. These variations may arise from differences in how effectively each model adheres to and aligns with the provided prompts. In future work, we plan to address these limitations by enhancing prompt adherence and alignment across different LLM architectures.

    \section*{Ethics Statement}
    Our data augmentation method leverages LLMs to generate data independently of the existing training dataset. However, it is important to note that the generated data may inherit social biases present in the pre-training corpus. Therefore, in practical applications, we recommend conducting manual reviews of the generated data to mitigate the risk of propagating biased information into the sequence labeling models.

    \bibliography{custom}

    \appendix

    \section*{Appendix}

    \section{Data Synthesis Prompts}

    \begin{figure*}[h!]
        \centering
        \includegraphics[width=0.95\textwidth]{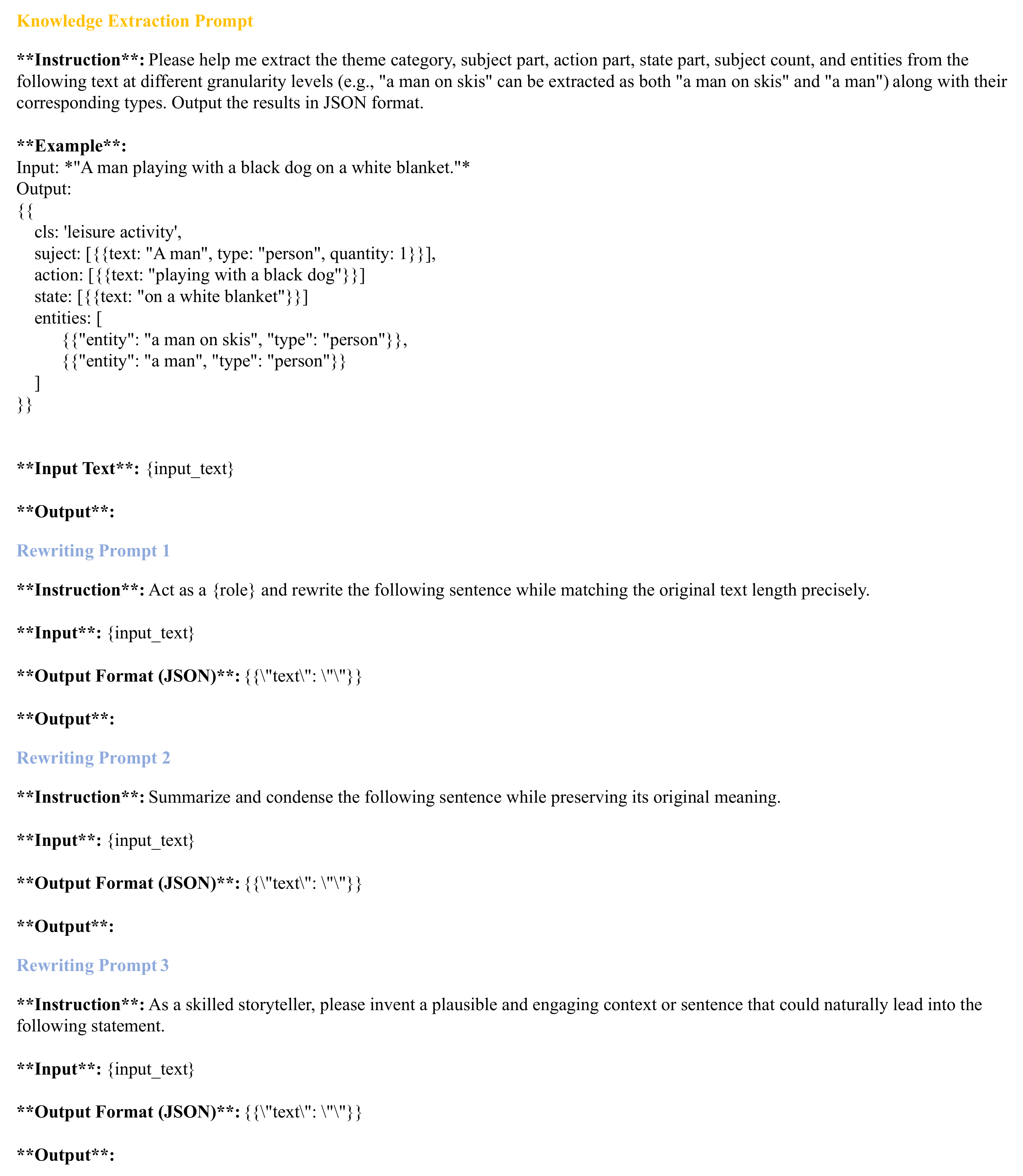}
        \caption{Examples of prompts used for data synthesis (Part 1).}
        \label{p1}
    \end{figure*}

    \begin{figure*}[h!]
        \centering
        \includegraphics[width=0.95\textwidth]{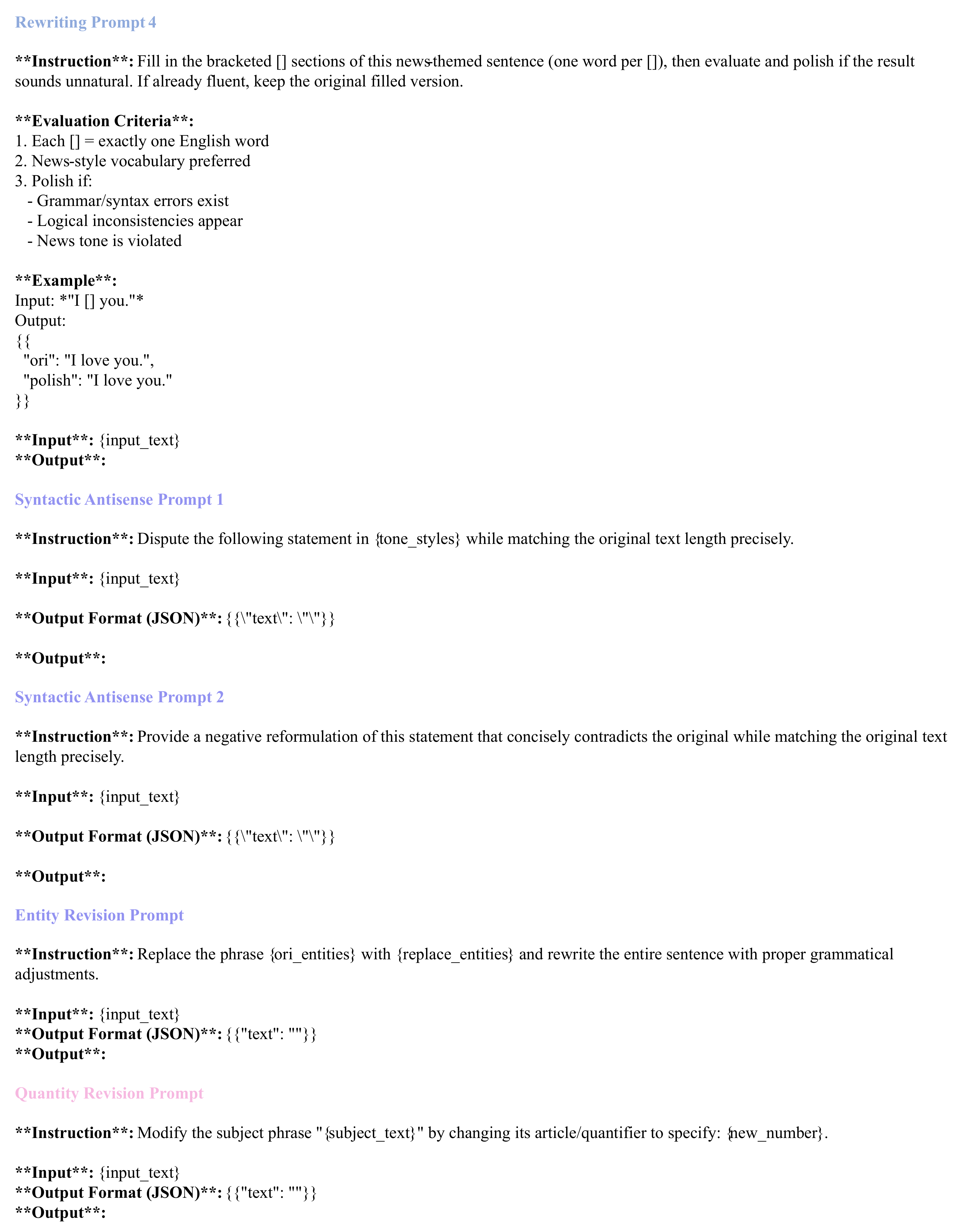}
        \caption{Examples of prompts used for data synthesis (Part 2).}
        \label{p2}
    \end{figure*}
    In this section, we provide the specifics of our prompts for knowledge extraction
    and integration, and data synthesis. The particular prompts are presented in
    Figure~\ref{p1} and~\ref{p2}.

    \begin{table*}
        [t!]
        \centering
        \resizebox{\textwidth}{!}{%
        \begin{tabular}{p{2.5cm}|m{5cm}<{\centering}|m{1.3cm}<{\centering}m{2.3cm}<{\centering}m{2.3cm}<{\centering}m{1.3cm}<{\centering}m{1.3cm}<{\centering}}
            \toprule \textbf{Model}                 & \textbf{Method}            & \textbf{AskU.} & \textbf{Mindsmall} & \textbf{SciDocsRR} & \textbf{StackO.} & \textbf{Avg.}  \\
            \midrule \multirow{4}{*}{BERT-base}     & SimCSE                     & 51.89          & 28.68              & 67.88              & {\ul 39.60}      & 47.01          \\
                                                    & PCL                        & 52.46          & 28.72              & 68.03              & \textbf{41.30}   & {\ul 47.63}    \\
                                                    & SynCSE (GPT-3.5 Turbo)*          & {\ul 52.61}    & \textbf{29.17}     & {\ul 68.46}        & 38.60            & 47.21          \\
                                                    & GCSE (ChatGLM3-6B)         & \textbf{52.62} & {\ul 28.79}        & \textbf{70.67}     & 39.53            & \textbf{47.90} \\
            \midrule \multirow{4}{*}{BERT-large}    & SimCSE                     & 53.10          & {\ul 29.59}        & {\ul 71.94}        & {\ul 40.68}      & {\ul 48.83}    \\
                                                    & PCL                        & 52.03          & 29.11              & 70.30              & \textbf{42.33}   & 48.44          \\
                                                    & SynCSE (GPT-3.5 Turbo)*          & {\ul 53.24}    & \textbf{30.09}     & 71.45              & 39.24            & 48.50          \\
                                                    & GCSE (ChatGLM3-6B)         & \textbf{53.40} & 29.43              & \textbf{73.04}     & 39.68            & \textbf{48.89} \\
            \midrule \multirow{5}{*}{RoBERTa-base}  & SimCSE$\dag\dag$           & 52.78          & {\ul 29.91}        & 65.96              & 39.25            & 46.95          \\
                                                    & CARDS$\dag\dag$            & 52.94          & 27.92              & 64.62              & \textbf{41.51}   & 46.75          \\
                                                    & PCL$\dag\dag$              & 51.85          & 27.92              & 64.70              & {\ul 41.18}      & 46.41          \\
                                                    & SynCSE (GPT-3.5 Turbo)$\dag\dag$ & {\ul 53.27}    & \textbf{30.29}     & {\ul 67.55}        & 39.39            & {\ul 47.63}    \\
                                                    & GCSE (ChatGLM3-6B)         & \textbf{53.44} & 29.35              & \textbf{67.89}     & 41.13            & \textbf{47.95} \\
            \midrule \multirow{5}{*}{RoBERTa-large} & SimCSE$\dag\dag$           & {\ul 55.10}    & 29.23              & 68.54              & {\ul 42.56}      & 48.86          \\
                                                    & CARDS$\dag\dag$            & 53.83          & 29.07              & 68.26              & \textbf{43.24}   & 48.60          \\
                                                    & PCL$\dag\dag$              & 53.43          & 28.56              & 66.06              & 41.54            & 47.40          \\
                                                    & SynCSE (GPT-3.5 Turbo)$\dag\dag$ & \textbf{55.48} & {\ul 30.27}        & {\ul 70.85}        & 40.00            & {\ul 49.15}    \\
                                                    & GCSE (ChatGLM3-6B)         & 54.05          & \textbf{30.30}     & \textbf{71.23}     & 41.65            & \textbf{49.31} \\
            \bottomrule
        \end{tabular}%
        }
        \caption{Comparison of Mean Average Precision (MAP) results on reranking
        tasks, where the value highlighted in bold is the best value, and the
        value underlined is the second-best value. ``$\dag\dag$'': results from
        \citet{DBLP:conf/emnlp/ZhangLH23}. ``*'': we reproduce the results with
        the officially released corpus from \citet{DBLP:conf/emnlp/ZhangLH23}.}
        \label{t2}
    \end{table*}

    \section{Reranking Tasks}
    \label{a_rk} To compare the ranking performance of our method on retrieval
    tasks, we evaluated the model using the MTEB benchmark \citep{muennighoff-etal-2023-mteb}
    with four reranking datasets: AskUbuntuDupQuestions \citep{lei-etal-2016-semi},
    MindSmallReranking \citep{wu-etal-2020-mind}, SciDocsRR \citep{cohan-etal-2020-specter}
    and StackOverflowDupQuestions \citep{DBLP:conf/sigsoft/Liu0LZ18}, and follow
    the same settings of \citet{DBLP:conf/emnlp/ZhangLH23} by using Mean Average
    Precision (MAP) as the metric.

    Table~\ref{t2} presents the MAP results of our approach and related
    baselines on the reranking benchmark, and all models are evaluated on the test
    sets of the reranking benchmark without using the training sets. The results
    indicate that various approaches exhibit varying performance on different
    datasets, which can be attributed to the distinct semantic distribution and evaluation
    scale of each dataset. Our GCSE outperforms SynCSE by 0.39\% in average MAP score
    and achieves the best results in all backbone models, demonstrating the
    efficacy of our approach in enhancing the precision of unsupervised ranking
    tasks.

    \section{Visualization of synthetic sample distribution}

    \begin{figure*}[!h]
        \centering
        \includegraphics[width=\textwidth]{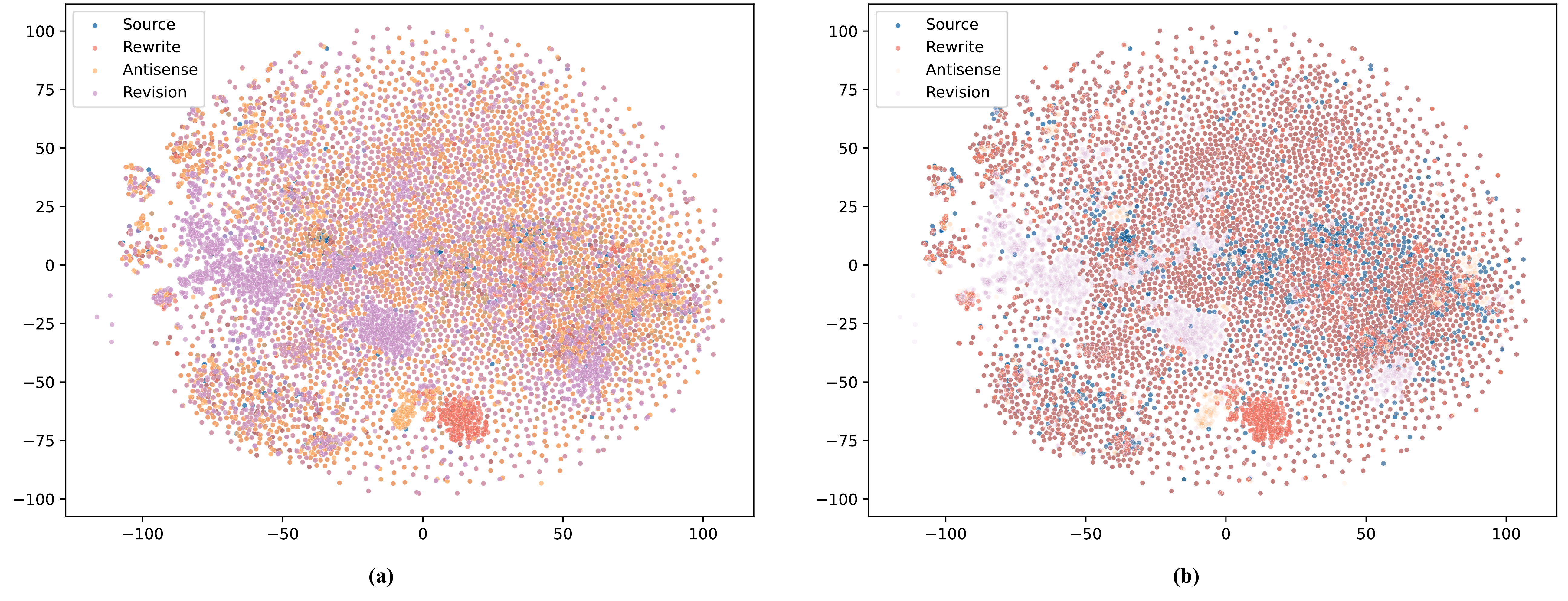}
        \caption{t-SNE visualization of the synthetic sample generated by
        ChatGLM3-6B, where the transparency of ``Antisense'' and ``Revision'' samples
        in subgraph (b) is reduced to 10\% for better observation.}
        \label{f7}
    \end{figure*}

    In this section, we use the supervised SimCSE model to generate sentence embeddings
    for the synthesized samples and utilize t-SNE to project the vectors into two-dimensional
    space for a visual analysis of the diversity. To facilitate observation, we group
    the synthesized samples into three categories: ``Rewrite'' refers to
    positive samples synthesized using ``Rewriting Prompt 1'' and ``Rewriting
    Prompt 2'' from Figure~\ref{p1}, while ``Antisense'' denotes the negative samples
    generated using ``Syntactic Antisense Prompt''. ``Revision'' denotes the
    negative samples generated using ``Entity Revision Prompt'', ``Quantity Revision
    Prompt'' and ``Rewriting Prompt 3'', which are related to knowledge
    modification. And ``Source'' indicates the original samples from the dataset.
    We randomly selected 5k ``Source'' samples and corresponding synthetic
    samples from our dataset for visualization, and the results are illustrated in
    Figure~\ref{f7}. We observe that ``Rewrite'' samples basically cover the
    spatial distribution of ``Source'' samples while expanding into the
    neighborhood space to some extent. ``Antisense'' and ``Revision'' samples further
    enhance the information density within the target semantic space. Comparing
    Figure~\ref{f7} (a) and (b), it can be observed that the ``Revision'' samples
    cover areas with sparse information, while their overall spatial
    distribution remains consistent with the semantic distribution of `Source'' samples.
    This indicates that the sample synthesis with knowledge effectively
    increases sample diversity within the semantic space.

    \section{Performance on Transfer Tasks}
    \label{a_trans}

    We also evaluate our GCSE following the same settings as SimCSE on seven transfer
    tasks: MR \citep{10.3115/1219840.1219855}, CR \citep{DBLP:conf/kdd/HuL04},
    SUBJ \citep{10.3115/1218955.1218990}, MPQA \citep{DBLP:journals/lre/WiebeWC05},
    SST2 \citep{socher-etal-2013-recursive}, TREC \citep{10.1145/345508.345577},
    and MRPC \citep{10.1145/345508.345577}. The results are shown in Table~\ref{t5},
    it can be observed that our GCSE (GPT-3.5 Turbo) achieves the best performance on all
    backbone models, outperforming second-best methods in average scores of 0.89\%
    with BERT-base, 0.79\% with BERT-large, 0.44\% with RoBERTa-base, and 0.40\%
    with RoBERTa-large, demonstrating the potential capability in downstream
    tasks.

        \begin{table*}
        [h!]
        \resizebox{\textwidth}{!}{%
        \centering
        \begin{tabular}{c|l|cccccccc}
            \toprule \textbf{Model}                 & \textbf{Method}               & \textbf{MR}    & \textbf{CR}    & \textbf{SUBJ}  & \textbf{MPQA}  & \textbf{SST2}  & \textbf{TREC}  & \textbf{MRPC}  & \textbf{Avg.}  \\
            \midrule \multirow{9}{*}{BERT-base}     & SimCSE$\spadesuit$            & 68.40          & 82.41          & 74.38          & 80.91          & 78.56          & 76.85          & 72.23          & 76.25          \\
                                                    & DiffCSE$\spadesuit$           & 72.28          & 84.43          & 76.47          & 83.90          & 80.54          & 80.59          & 71.23          & 78.49          \\
                                                    & PCL$\spadesuit$               & 72.84          & 83.81          & 76.52          & 83.06          & 79.32          & 80.01          & 73.38          & 78.42          \\
                                                    & RankCSE$\spadesuit$           & 75.66          & 86.27          & 77.81          & 84.74          & 81.10          & 81.80          & 75.13          & 80.36          \\
                                                    & MultiCSR (GPT-3.5 Turbo)$\clubsuit$ & 82.70          & 88.15          & 94.97          & 90.08          & 86.87          & \textbf{87.70}          & 75.46          & 86.56          \\
                                                    & SynCSE (GPT-3.5 Turbo)*             & 83.34          & 88.80          & 93.88          & 90.39    & 88.96    & 83.60          & 75.94          & 86.42          \\
                                                    & \textbf{GCSE (ChatGLM3-6B)}   & \textbf{84.79} & 90.03          & 94.35          & 89.92          & 88.37          & 85.60          & 75.71          & 86.97          \\
                                                    & \textbf{GCSE (GLM4-9B-Chat)}       & 84.53          & 89.96          & {\ul 95.01}          & 89.97          & 88.67          & 86.21          & 76.01          & 87.19          \\
                                                    & \textbf{GCSE (Qwen2.5-32B-Instruct)}   & 83.94          & 89.65          & 94.71          & 90.31          & 88.25          & 86.00          & \textbf{76.46} & 87.05          \\
                                                    & \textbf{GCSE (GPT-3.5 Turbo)} & {\ul 84.71}          & \textbf{90.18} & 94.32          & {\ul 90.61}          & \textbf{89.53} & 86.09          & 76.22          & {\ul 87.38}          \\
                                                    & \textbf{GCSE (Deepseek-V3-0324)}   & 84.66          & {\ul 90.07}          & \textbf{95.02} & \textbf{90.62} & {\ul 89.16}          & {\ul 86.37} & {\ul 76.28}          & \textbf{87.45} \\
            \midrule \multirow{7}{*}{BERT-large}    & SimCSE$\spadesuit$            & 70.88          & 84.16          & 76.43          & 84.50          & 79.76          & 79.26          & 73.88          & 78.41          \\
                                                    & PCL$\spadesuit$               & 74.87          & 86.11          & 78.29          & 85.65          & 80.52          & 81.62          & 73.94          & 80.14          \\
                                                    & RankCSE$\spadesuit$           & 75.48          & 86.50          & 78.60          & 85.45          & 81.09          & 81.58          & 75.53          & 80.60          \\
                                                    & SynCSE (GPT-3.5 Turbo)*             & 85.78    & 90.47    & 94.77          & 90.41    & 90.50    & 89.00          & 75.77 & 88.10    \\
                                                    & \textbf{GCSE (ChatGLM3-6B)}   & 86.08          & 90.54          & 95.00          & {\ul 90.63}          & 91.21          & {\ul 89.60}          & 75.71          & 88.40          \\
                                                    & \textbf{GCSE (GLM4-9B-Chat)}       & 86.33          & 90.26          & 95.08          & \textbf{90.65} & {\ul 92.13}          & \textbf{92.08} & 75.63          & \textbf{88.88} \\
                                                    & \textbf{GCSE (Qwen2.5-32B-Instruct)}   & {\ul 86.35}          & {\ul 90.73}          & \textbf{95.18} & 90.60          & 91.93          & 87.80          & {\ul 76.12}          & 88.39          \\
                                                    & \textbf{GCSE (GPT-3.5 Turbo)} & 85.77          & \textbf{90.88} & 94.35          & 90.09          & \textbf{92.91} & 88.91          & 75.12          & 88.29          \\
                                                    & \textbf{GCSE (Deepseek-V3-0324)}   & \textbf{86.46} & 90.46          & {\ul 95.06}          & 90.49          & 91.93          & 87.80          & \textbf{76.87} & {\ul 88.44}          \\
            \midrule \multirow{9}{*}{RoBERTa-base}  & SimCSE$\spadesuit$            & 70.16          & 81.77          & 73.24          & 81.36          & 80.65          & 80.22          & 68.56          & 76.57          \\
                                                    & DiffCSE$\spadesuit$           & 70.05          & 83.43          & 75.49          & 82.81          & 82.12          & 82.38          & 71.19          & 78.21          \\
                                                    & PCL$\spadesuit$               & 71.13          & 82.38          & 75.40          & 83.07          & 81.98          & 81.63          & 69.72          & 77.90          \\
                                                    & RankCSE$\spadesuit$           & 73.20          & 85.95          & 77.17          & 84.82          & 82.58          & 83.08          & 71.88          & 79.81          \\
                                                    & MultiCSR (GPT-3.5 Turbo)$\clubsuit$ & 84.70          & 90.69          & 94.40    & 89.38          & 89.42          & \textbf{89.62} & \textbf{77.01} & 87.89    \\
                                                    & SynCSE (GPT-3.5 Turbo)$\dag\dag$    & 85.47    & 91.44    & 92.53          & {\ul 89.67}    & 90.94          & 81.60          & 76.06          & 86.82          \\
                                                    & \textbf{GCSE (ChatGLM3-6B)}   & {\ul 86.79}          & {\ul 92.03}          & 94.35          & 89.92 & 92.37          & 85.60          & 75.71          & 88.11          \\
                                                    & \textbf{GCSE (GLM4-9B-Chat)}       & \textbf{86.91} & \textbf{92.14} & \textbf{94.62} & 89.76          & \textbf{92.60} & 86.21          & 76.17          & \textbf{88.34} \\
                                                    & \textbf{GCSE (Qwen2.5-32B-Instruct)}   & 86.32          & 91.58          & 94.37          & {\ul 90.04}          & 92.42          & 84.00          & 76.12          & 87.84          \\
                                                    & \textbf{GCSE (GPT-3.5 Turbo)} & 86.66          & 91.57          & {\ul 94.44}          & \textbf{90.82}          & {\ul 92.45}          & 84.93          & {\ul 76.18} & 88.15          \\
                                                    & \textbf{GCSE (Deepseek-V3-0324)}   & 86.42          & 91.56          & 94.41          & 89.23          & 92.18          & {\ul 87.52} & 76.13          & {\ul 88.21}          \\
            \midrule \multirow{7}{*}{RoBERTa-large} & SimCSE$\spadesuit$            & 72.86          & 83.99          & 75.62          & 84.77          & 81.80          & 81.98          & 71.26          & 78.90          \\
                                                    & PCL$\spadesuit$               & 74.08          & 84.36          & 76.42          & 85.49          & 81.76          & 82.79          & 71.51          & 79.49          \\
                                                    & RankCSE$\spadesuit$           & 73.20          & 85.83          & 78.00          & 85.63          & 82.67          & 84.19          & 73.64          & 80.45          \\
                                                    & SynCSE (GPT-3.5 Turbo)$\dag\dag$    & 87.24          & 92.16 & 93.75          & {\ul 90.81} & 91.87          & 84.00          & \textbf{76.29} & 88.02          \\
                                                    & \textbf{GCSE (ChatGLM3-6B)}   & {\ul 87.60}          & \textbf{92.43} & 94.66          & 90.36          & \textbf{92.37} & 88.80          & {\ul 75.30} & {\ul 88.79}          \\
                                                    & \textbf{GCSE (GLM4-9B-Chat)}       & 85.55          & 90.39          & {\ul 94.70}          & 90.37          & 90.32          & \textbf{92.65} & 73.19          & 88.17          \\
                                                    & \textbf{GCSE (Qwen2.5-32B-Instruct)}   & \textbf{87.73} & {\ul 92.18}          & \textbf{94.72} & 90.68          & {\ul 92.26}          & {\ul 90.00}          & 74.20          & \textbf{88.82} \\
                                                    & \textbf{GCSE (GPT-3.5 Turbo)} & 87.12          & 91.98          & 94.01          & 90.71          & 92.25          & 88.75          & 74.55          & 88.48          \\
                                                    & \textbf{GCSE (Deepseek-V3-0324)}   & \textbf{87.73} & {\ul 92.18}          & 94.29          & \textbf{90.95} & 92.15          & 88.80          & 73.28          & 88.48          \\
            \bottomrule
        \end{tabular}%
        }
        \caption{Comparison of different sentence embedding models accuracy on
        transfer tasks. ``$\spadesuit$'': results from
        \citet{DBLP:conf/acl/LiuLWWWX0C023}, ``$\clubsuit$'': results from
        \citet{DBLP:conf/naacl/WangLCSB24}, ``$\dag\dag$'': results from
        \citet{DBLP:conf/emnlp/ZhangLH23}. ``*'': we reproduce the results with
        the officially released corpus from \citet{DBLP:conf/emnlp/ZhangLH23}.}
        \label{t5}
    \end{table*}

    \section{Case Studies}

    \begin{table*}
        [h!]
        \resizebox{\textwidth}{!}{%
        \begin{tabular}{ccccccc}
            \toprule Premise                                                                                                                   & Hypothesis                                                                                                     & \textbf{Gold} & \textbf{SimCSE}       & \textbf{RankCSE} & \textbf{SynCSE}    & \textbf{GCSE}         \\
            \midrule A woman is cooking \colorbox[RGB]{143,170,220}{eggs}.                                                                     & A woman is cooking \colorbox[RGB]{143,170,220}{something}.                                                     & 3.00          & 4.37 (1.372)          & 4.23 (1.320)     & {\ul 3.66 (0.662)} & \textbf{3.24 (0.236)} \\
            \colorbox[RGB]{255,217,109}{Two} little girls are talking on the phone.                                                            & \colorbox[RGB]{255,217,109}{A} little girl is walking down the street.                                         & 0.50          & 3.38 (2.881)          & 3.64 (3.139)     & {\ul 1.97 (1.468)} & \textbf{1.85 (1.351)} \\
            A chef is preparing \colorbox[RGB]{255,217,109}{some} \colorbox[RGB]{143,170,220}{food}.                                           & \colorbox[RGB]{255,217,109}{A} chef prepared \colorbox[RGB]{255,217,109}{a} \colorbox[RGB]{143,170,220}{meal}. & 4.00          & \textbf{4.27 (0.270)} & 4.59 (0.588)     & 4.56 (0.561)       & {\ul 4.41 (0.408)}    \\
            \colorbox[RGB]{255,217,109}{Five} kittens are eating out of \colorbox[RGB]{255,217,109}{five} \colorbox[RGB]{143,170,220}{dishes}. & Kittens are eating \colorbox[RGB]{143,170,220}{food} on trays.                                                 & 2.75          & 3.81 (1.056)          & 3.71 (0.957)     & {\ul 3.28 (0.535)} & \textbf{3.12 (0.373)} \\
            A woman is cutting \colorbox[RGB]{255,217,109}{some} \colorbox[RGB]{143,170,220}{herbs}.                                           & A woman is chopping \colorbox[RGB]{143,170,220}{cilantro}.                                                     & 2.80          & 3.58 (0.777)          & 3.58 (0.967)     & {\ul 3.11 (0.313)} & \textbf{2.61 (0.185)} \\
            \bottomrule
        \end{tabular}%
        }
        \caption{Case studies on model prediction similarity with gold labels in
        the STS-Benchmark development set, where Gold represents the label score
        of the sentence pair (ranging from zero to five). The similarity scores
        of all models are multiplied by a coefficient of five for better
        comparison, and the value in parentheses denotes the RMS error between the
        predicted score and the label. Words highlighted in blue denote the
        entity alteration in the sentence-pair, whereas words in yellow indicate
        the quantities that change inside the sentence-pair.}
        \label{t6}
    \end{table*}

    To further verify the improvement in our method's awareness of entity and quantity,
    we selected five sample sets from the STS-Benchmark development set that
    explicitly contained alterations in entity or quantity within the sentence-pair,
    and presented the prediction cosine-similarity scores of GCSE (ChatGLM3-6B) and related methodologies
    with the backbone of BERT-base in Table~\ref{t6}. We can observe from the
    results that the prediction score of our model achieves the minimum root-mean-square (RMS)
    error compared to the label in most cases, which indicates that our model has
    a stronger capacity to distinguish information.

    \section{Ablation Studies of Gaussian-decayed and Few-shot Samples}
    \label{a_gau}

    \begin{table*}
        [h!]
        \resizebox{\textwidth}{!}{%
        \begin{tabular}{lcccccccc}
            \toprule \textbf{Method}                 & \textbf{STS-12} & \textbf{STS-13} & \textbf{STS-14} & \textbf{STS-15} & \textbf{STS-16} & \textbf{STS-B} & \textbf{SICK-R} & \textbf{Avg.}  \\
            \midrule SynCSE (GPT-3.5 Turbo)*               & 75.86           & 82.19           & 78.71           & 85.63           & \textbf{81.11}  & 82.35          & \textbf{78.79}  & 80.66          \\
            \hspace{1em} w sampled                   & 75.48           & 85.60           & 78.76           & 84.78           & 80.38           & 82.12          & 76.46           & 80.51          \\
            \hspace{1em} w sampled \& G.D.           & 75.71           & 85.24           & 79.09           & 85.15           & 80.82           & 82.68          & 77.54           & 80.89          \\
            \hspace{1em} w G.D.                      & \textbf{75.89}  & 85.26           & 79.24           & 85.67           & 80.79           & 82.63          & 78.19           & \textbf{81.10} \\
            \hspace{1em} w sampled \& domain \& G.D. & 75.88           & \textbf{86.02}  & \textbf{79.46}  & \textbf{86.10}  & 80.27           & \textbf{82.87} & 76.91           & 81.07          \\
            \bottomrule
        \end{tabular}%
        }
        \caption{Ablation studies of sample size and the Gaussian-decayed
        function by utilizing SynCSE. ``*'': we reproduce the results with the
        officially released corpus from \citet{DBLP:conf/emnlp/ZhangLH23}.}
        \label{t7}
    \end{table*}

    We employ the Gaussian-decayed function on SynCSE and sample SynCSE training
    data with a sample size the same as our synthetic data to evaluate the efficacy
    of the proposed Gaussian-decayed function and our domain-oriented selection strategy in
    the ablation experiment. The data sample size is 64k, and the weight of $\sigma$
    in $G(\cdot)$ is assigned the same value as specified in Section~\ref{exp_setup}.
    The results of various policies implemented in SynCSE are presented in Table~\ref{t7}.
    ``w sampled'' denotes the utilization of purely the sampled data in SynCSE,
    and a performance decrease can be observed when training on a reduced number
    of samples without extra configurations. ``w sampled \& G.D.'' denotes the
    additional incorporation of $G(\cdot)$ based on ``w sampled''. ``w G.D.''
    indicates the results by training on the full dataset utilizing $G(\cdot)$. In
    both configurations, the average performance outperforms the vanilla model, illustrating
    the module's efficacy. “w sampled \& domain \& G.D.” denotes the concurrent
    utilization of sample data, domain data, and $G( \cdot)$, with a sample size
    of 48k for the SynCSE dataset and 16k for the synthesized domain dataset.
    The results reveal that "w sampled \& domain \& G.D." attains the second-best
    performance, suggesting that incorporating domain data can decrease the
    required training samples while enhancing model efficacy.

    \section{Unsupervised Sentence Embedding on LLM}

    \begin{table*}
        [h!]
        \centering
        \begin{tabular}{p{5.5cm}c|p{5.5cm}c}
            \toprule Model                                      & Avg.                                          & Model            & Avg.           \\
            \midrule \multicolumn{2}{c|}{\textit{Unsupervised}} & \multicolumn{2}{c}{\textit{Data Augmentation}} \\
            \midrule Llama3.2-3B-Instruct LoRA                           & 71.34                                         & Llama3.2-3B-Instruct LoRA & 78.26          \\
            Llama-3-8B-Instruct LoRA                                     & \textbf{72.73}                                & Llama-3-8B-Instruct LoRA  & 78.24          \\
            ChatGLM3-6B LoRA                                    & 69.38                                         & ChatGLM3-6B LoRA & 79.04          \\
            GLM4-9B-Chat LoRA                                        & 71.77                                         & GLM4-9B-Chat LoRA     & \textbf{79.52} \\
            Qwen2.5-14B-Instruct LoRA                                    & 68.49                                         & Qwen2.5-14B-Instruct LoRA & 78.02          \\
            \bottomrule
        \end{tabular}
        \caption{Performance comparison of different LLMs on STS tasks, where results
        of ``Unsupervised'' refers to models trained on the same unsupervised settings
        as \citet{gao2021simcse}, and ``Data Augmentation'' refers to models
        trained with the synthetic data generated by ChatGLM3-6B.}
        \label{t8}
    \end{table*}

    In this section, we utilize contrastive learning on multiple LLMs to evaluate
    the alignment of LLM-generated similarities with the gold labels and the effectiveness
    of our data augmentation strategy. We use Llama3.2-3B-Instruct \citep{DBLP:journals/corr/abs-2407-21783},
    Llama3-8B-Instruct \citep{DBLP:journals/corr/abs-2407-21783}, ChatGLM3-6B \citep{glm2024chatglm},
    GLM4-9B-Chat \citep{glm2024chatglm} and Qwen2.5-14B-Instruct \citep{qwen2.5, qwen2} with a
    low-rank adapter (LoRA) layer for training. The sentence embedding vectors are
    obtained from the output hidden states of the last position, which is
    followed by the method of pretended chain of thought (Pretended CoT) \citep{10.1007/978-981-97-5669-8_5}.
    We may derive two major conclusions from the results in Table~\ref{t8}: (1) In
    conventional unsupervised settings, decoder-based LLMs have no significant
    performance advantage over encoder-based PLMs for sentence representation
    learning tasks. The model performance does not increase significantly with
    the increase of the number of model parameters. To reduce expenses, we
    assert that fully leveraging the capabilities of LLMs for distilling smaller
    models is the better option. (2) The application of our data augmentation technique
    to sentence representation learning tasks in LLMs significantly enhances performance
    relative to the ``Unsupervised'' settings, which further proves the
    applicability and efficacy of our strategy.

    \section{Visualization of Prediction Scores and Gradient Comparisons}\label{a_vis_gau}

    \begin{figure*}[h!]
        \centering
        \includegraphics[width=\textwidth]{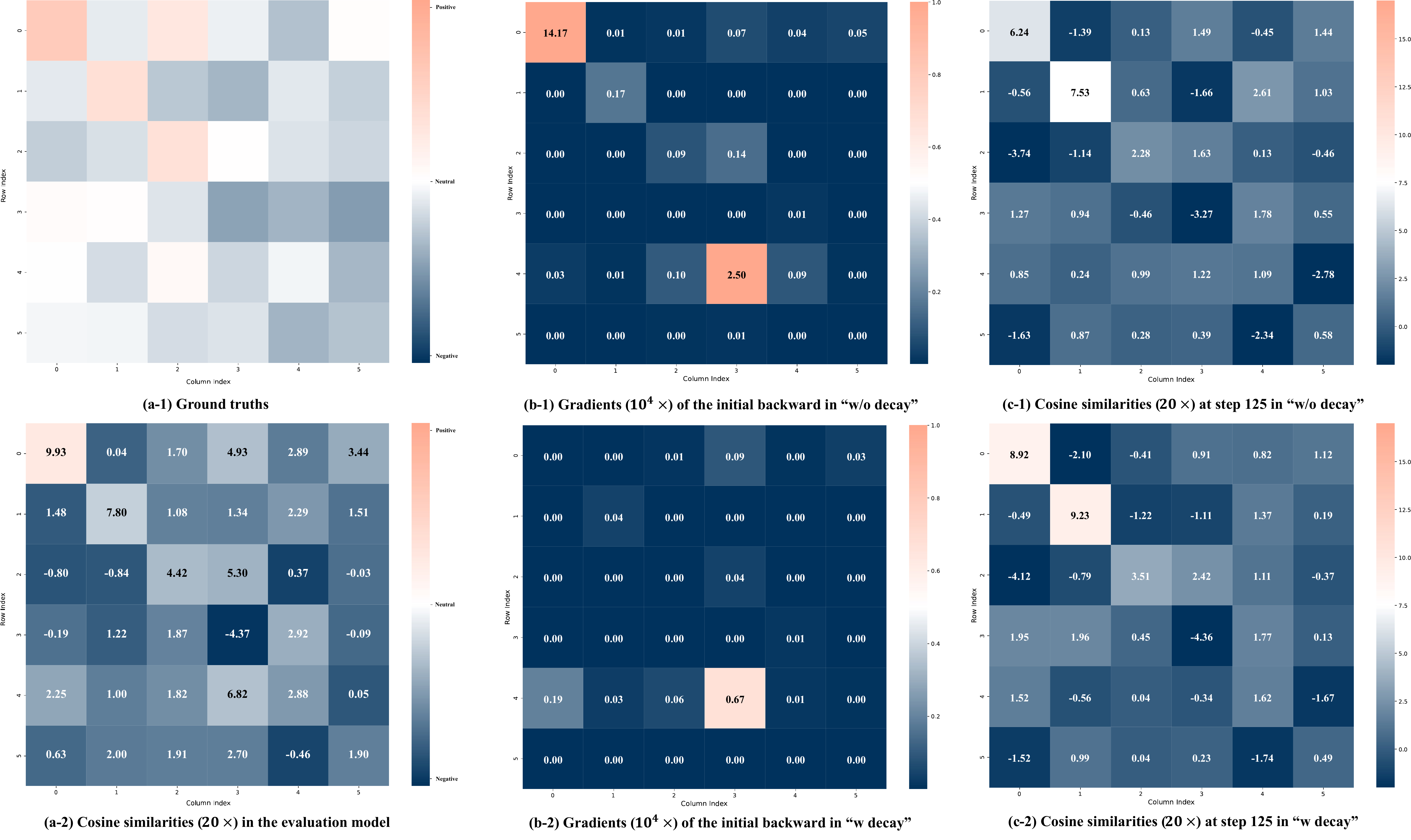}
        \caption{Heatmap visualization of the prediction scores and gradients.}
        \label{f8}
    \end{figure*}

    To further analyze the effectiveness of the Gaussian-decayed function in
    mitigating the impact of false negative noise, we visualized the changes in predicted
    scores and gradients during the training process using heatmaps. In the
    training procedure of GCSE, each input consists of a source sample, its
    corresponding positive sample, and a hard negative sample. We visualize the
    cosine similarity scores and gradient heatmaps for negative samples within a
    batch in Figure~\ref{f8}. Each cell of a heatmap represents the relationship
    between the source sample and the negative sample, and the diagonal cells
    highlight the relationships between source samples and their hard negatives.
    Since synthetic samples lack manual annotations, we use supervised SimCSE
    models \citep{gao2021simcse} based on different backbones to compute their similarity
    scores as the ground truth. We normalized the output scores of each model
    with min-max scaling and averaged them as the final scores to address
    distributional differences across models, and the results are shown in Figure~\ref{f8}
    (a-1). It can be observed that several hard negatives on the diagonal
    display scores biased towards positive similarity, indicating the presence of
    false negative noise. In the framework of contrastive learning, when optimized
    using standard contrastive loss, these hard negatives are positioned further
    from the source samples in the semantic space, negatively impacting the model's
    representational capacity. Figure~\ref{f8} (a-2) displays the normalized
    cosine similarity scores of hard negatives in the initial step as calculated
    by the evaluation model in GCSE. The initial score distribution of hard negatives
    shows a strong correlation with the ground truth, suggesting that these
    scores could efficiently guide GCSE in gradient correction.

    Figures~\ref{f8} (b-1) and (b-2) present the backward gradient values of the
    model trained without and with the Gaussian-decayed function, respectively.
    For better visualization, all gradient values are amplified by $10^{4}$, and
    all similarities are amplified by $20$ by the temperature. By comparing the gradients
    of hard negative samples in these two figures, it can be observed that the
    gradient values on false hard negatives are significantly smaller when the
    Gaussian-decayed function is applied. Additionally, Figures~\ref{f8} (c-1)
    and (c-2) present a comparison of cosine similarity scores after 125
    training steps with and without the Gaussian-decayed function. The scores for
    false hard negatives are significantly higher when the Gaussian-decayed function
    is employed, while the true hard negatives had lower scores. The overall score
    distribution aligns more accurately with the ground truth, and these results
    demonstrate that the Gaussian-decayed function effectively prevents false
    negatives from being pushed farther away from source samples in the semantic
    space, thereby validating its effectiveness in mitigating noise and improving
    model performance.

    \section{Ablation analysis of filtering thresholds}\label{a_filtering}

    \begin{figure}[!h]
        \centering
        \includegraphics[width=\columnwidth]{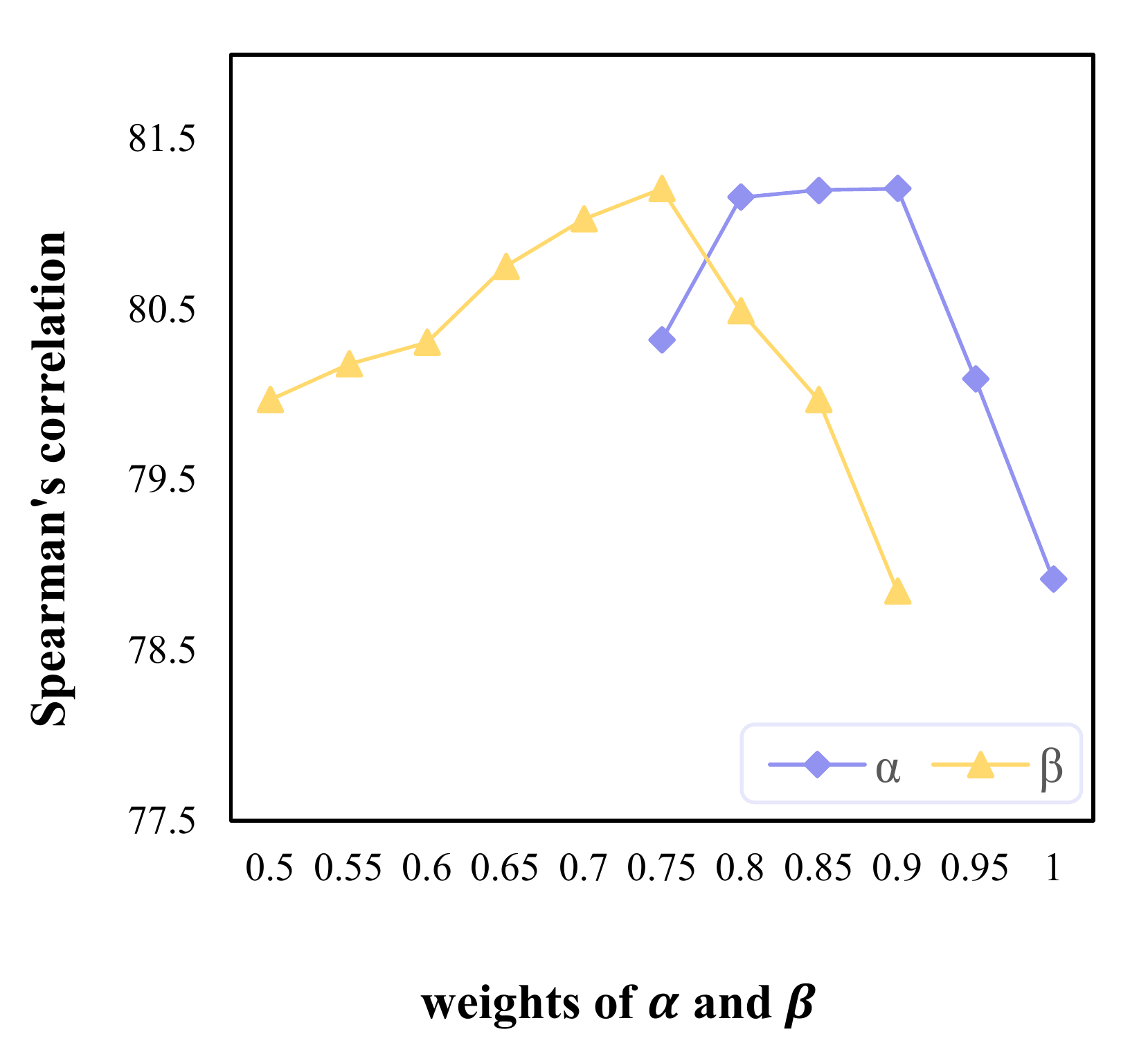}
        \caption{Spearman's correlation against the weight of $\alpha$ and
        $\beta$ on the STS tasks. When adjusting the weight of one parameter,
        the other parameter is fixed at its default value as specified in the
        experimental settings.}
        \label{f9}
    \end{figure}

    To study the impact of different filtering thresholds, we evaluate the
    performance on the backbone of the BERT-base, and the results are shown in Figure~\ref{f9}.
    When $\alpha > 0.9$, the model's performance declines significantly,
    primarily because the high threshold filters out too many samples, heavily reducing
    the number of positive samples. In the range $\alpha \in [0.8,0.9]$, performance
    degradation is observed due to noise introduced by false positive samples.
    Similarly, when $\alpha < 0.8$, the model suffers from a performance drop
    caused by an excessive number of false positives being included in the
    training process. The threshold for $\beta$ demonstrates a noticeable impact
    on model performance when it deviates from 0.75. Specifically, when $\beta >
    0.75$, the model's performance declines significantly due to the inclusion
    of excessive false negative noise, which severely affects the model performance.
    Conversely, when $\beta < 0.75$, the selected negative samples become easier
    for the model to distinguish, providing limited benefit for enhancing its representation
    learning capacity. The results highlight the influence of filtering
    thresholds on sample quality and distribution.

    \section{Score Normalization Methodology}
    \label{ap_snm}

    In this work, the labels in datasets are normalized with standard min-max
    normalization. To address the discrepancy in score distributions among different
    models, we applied a variant min-max normalization method to align their
    predicted scores. For each label $l \in [0, \text{MAX}]$, we collect all
    predicted scores with $l = 0$ as list $C_{0}$, and all predicted scores with
    $l = \text{MAX}$ as list $C_{1}$. Specifically, we computed the median prediction
    scores for $C_{0}$ and $C_{1}$ as $min_{p}= \operatorname{median}(C_{0})$ and
    $max_{p}= \operatorname{median}(C_{1})$, respectively. The use of medians,
    rather than the minimum predicted score for $C_{0}$ or the maximum predicted
    score for $C_{1}$, avoids reliance on outlier values that may disproportionately
    skew the normalization, ensuring a more balanced score distribution. For a given
    score $s$, the normalized score $s^{\prime}$ is calculated as:

    \begin{equation}
        s^{\prime}= \text{clip}\left(\frac{s - min_{p}}{max_{p}- min_{p}}, 0, 1 \right
        ),
    \end{equation}

    where the function $\text{clip}(x, 0, 1)$ ensures the normalized score is bounded
    within $[0, 1]$. This method adjusts the score range to maintain consistency
    across models while preserving relative score differences.
\end{document}